\documentclass[10pt, a4paper]{article}
\usepackage{comment}
\usepackage[UTF8]{ctex} 
\usepackage{booktabs}

\usepackage[]{lrec2026} 
\usepackage{xcolor}

\title{Towards a resource for multilingual lexicons: an MT assisted and human-in-the-loop multilingual parallel corpus with multi-word expression annotation}

\name{Lifeng Han$^{1,2}$, Najet Hadj Mohamed$^{3}$, Malak Rassem$^{4}$ \\ \textbf{Gareth J. F. Jones}$^5$,  \textbf{Alan F. Smeaton}$^6$, \textbf{Goran Nenadic}$^{7}$} 

\address{ $^1$Biomedical Data Sciences, Leiden University Medical Center, NL \\ $^2$Leiden Institute of Advanced Computer Science (LIACS), Leiden University, NL\\
$^3$University of Tours, France, and \\  Arabic Natural Language Processing Research Group,  University of Sfax, Tunisia\\
$^4$Institute for Natural Language Processing (IMS),  University of Stuttgart, Germany\\
$^5$ADAPT Research Centre, Dublin City University, Dublin, Ireland\\
$^6$Insight Centre for Data Analytics, Dublin City University, Dublin, Ireland\\
$^7$University of Manchester, Manchester M13 9PL, The United Kingdom\\
         l.han@liacs.leidenuniv.nl l.han@lumc.nl\\}

\abstract{
In this work, we introduce the construction of a machine translation (MT) assisted and human-in-the-loop multilingual parallel corpus with annotations of multi-word expressions (MWEs), named AlphaMWE. The MWEs include verbal MWEs (vMWEs) defined in the PARSEME shared task that have a verb as the head of the studied terms. The annotated vMWEs are also bilingually and multilingually aligned manually. 
The languages covered include Arabic, Chinese, English, German, Italian, and Polish, of which, the Arabic corpus includes both standard and dialectal variations from Egypt and Tunisia.  
Our original English corpus is extracted from the PARSEME shared task in 2018. We performed machine translation of this source corpus followed by human post-editing and annotation of target MWEs. 
Strict quality control was applied for error limitation, i.e., each MT output sentence received first manual post-editing and annotation plus a second manual quality rechecking till annotators' consensus is reached.
One of our findings during corpora preparation is that accurate translation of MWEs presents challenges to MT systems, as reflected by the outcomes of human-in-the-loop metric HOPE. 
To facilitate further MT research, we present a categorisation of the error types encountered by MT systems in performing MWE-related translation.
To acquire a broader view of MT issues, we selected four popular state-of-the-art MT systems for comparison, namely Microsoft Bing Translator, GoogleMT, Baidu Fanyi, and DeepL MT.
Because of the noise removal, translation post-editing, and MWE annotation by human professionals, we believe the AlphaMWE data set will be an asset for both monolingual and cross-lingual research, such as multi-word term lexicography, MT, and information extraction. 
 \\ \newline  }

\begin{document}

\maketitleabstract


\section{Introduction}
\label{intro}
\footnote{Accepted manuscript by Language Resources and Evaluation Journal  \url{https://link.springer.com/journal/10579}}
Multi-word Expressions (MWEs) have long been of interest to both natural language processing (NLP)  researchers and linguists \cite{Sag2002MWE}, \cite{mwe2017survey}, \cite{2020phraseological}. The  automatic processing of MWEs has posed significant challenges for some tasks in computational linguistics  (CL), such as word sense disambiguation (WSD), parsing, and (automated) translation \cite{lambert2005mwe}, \cite{Bouamor2012IdentifyingBM}, \cite{Skadina2016MultiwordEI}, \cite{Li2016neuralname}, \cite{Han2020multimwe}. This is caused by  both the variety and the richness of MWEs as they are used in language.

Various definitions of MWEs have included both syntactic structure and semantic viewpoints from different researchers covering syntactic anomalies, non-compositionality, non-substitutability and ambiguity \cite{mwe2017survey}. For instance, \cite{mwe2010handbook} define MWEs as ``lexical items that: (i) can be decomposed into multiple lexemes; and (ii) display lexical, syntactic, semantic, pragmatic and/or statistical idiomaticity". 
%
The annual MWE workshop organized by SIGLEX-MWE group (Special Interest Group in Lexicon), from the Association for Computational Linguistics (ACL)\footnote{\url{http://multi-word.sourceforge.net/PHITE.php}}, has focused on the interests on MWEs in different fields of NLP research.
A scientific shared task on MWE discovery and identification has been held for 3 years to date (2017, 2018, and 2020)\cite{mwe2017sharedtask,PARSEME2018task,mwe-2020-joint}, where monolingual corpus with verbal MWEs annotated was offered for challenges across different languages, mainly European ones. Another related workshop series that had been held every two years were from EUROPHRAS `` Workshop on Multi-word Units in Machine Translation and Translation Technologies" since 2013\footnote{http://www.lexytrad.es/europhras2019/mumttt-2019-2/}.
However, as noted
by other NLP researchers for example in \cite{mwe2017survey},  there are very few bilingual or even multilingual parallel corpora with MWE annotations available
for cross-lingual NLP research and for downstream applications such as machine translation (MT) \cite{Google2016MultilingualNMT,zaninello-birch-2020-multiword}.

With regard to
MWE research, verbal MWEs (vMWEs) are 
a mature category that have received attention from many researchers \cite{maldonadoHanMoreau2017detection,Moreau2018SemanticRO}. 
Verbal MWEs have a verb as the head of the studied term and function as verbal phrases, such as ``\textit{kick} the bucket", ``\textit{cutting} capers" and ``\textit{go} to one's head".
In this work, we present the construction of a multilingual corpus with vMWEs annotation, including English-Chinese, English-German, English-Polish, English-Italian, and English-Arabic language pairs, with a main focus on English-Chinese and English-Arabic language pairs. We started with the same source monolingual corpus  in English with its vMWE tags from the shared task affiliated with the SIGLEX-MWE workshop in 2018 \cite{mwe2018english,PARSEME2018task}. Several state-of-the-art (SOTA) MT models were used to perform an automated translation, and then human post-editing and annotation for the target languages was conducted with cross-validation to ensure the quality, i.e., with each sentence receiving post-editing and rechecking by at least two people. For dialectal Arabic, we carried out the translation from scratch without MT in the loop because the current state-of-the-art systems don't have such options and it is easier to translate manually than post-editing.

In order to get a deeper insight into the difficulties of processing MWEs we carried out a categorisation of the errors 
made by
MT models 
when processing MWEs, especially using English-Chinese language pairs. 
We also produced quantitative analysis scores using the human post-editing metric HOPE on the English-Arabic language.
From these, we conclude that  current state-of-the-art MT models are far from reaching parity with humans  
in terms of translation performance, especially on idiomatic MWEs, and even for sentence level translation, although some researchers sometimes claim otherwise \cite{google2016bridging,microsoft2018AchievingHP}.

The rest of this paper is organised as follows. In the next section (\ref{sec_related_work}), we present 
related work and then 
detail the corpus preparation stages including selection of MT models and the resulting AlphaMWE (Section \ref{corpus_construction_section}). We then look at various  issues that MT has with MWEs (Section \ref{sec_MT_issues}), followed by a quantitative evaluation section using post-editing metric HOPE \cite{Gladkoff_Han_HOPE} on English-Arabic language pair as a case study (Section \ref{sec_en-ar-HOPE}). Subsequently, we present a broader discussion section (\ref{sec_Discussion}). This analysis and discussion, along with the public release of the corpora as a resource for the community, is the main contribution of the paper \footnote{We adopt the same license as the original PARSEME English dataset, i.e. CC-BY-SA 4.0}. 
Finally, we conclude the paper with a plan for future work in Section \ref{sec_conclude}. This is an extended work of our earlier MWE-workshop paper \cite{han-etal-2020-alphamwe}, where we add two more language pairs English-Italian and English-Arabic \cite{mohamed2023alphamwe_Arabic} for the corpora, add Polish and German sub-sections on error analysis from MT systems, quantitative evaluation, extensive discussion and analysis sections on MWEs related MT issues. 

\section{Related Work}
\label{sec_related_work}

There is a number of existing studies that focus on the creation of \textit{monolingual} corpora with vMWE annotations, such as the PARSEME shared task corpora \cite{parseme2017shared,PARSEME2018task}.  The 2020 edition of this task  covers 14 languages including Chinese, Hindi, and Turkish as non-European languages. \cite{vincze-csirik-2010-hungarian} prepared a manual annotation of Hungarian corpus with Light Verb Constructions (LVCs) and indicated their usability for machine translation and information extraction.
Some work from monolingual English corpora includes the ``MWE-aware English Dependency Corpus" from the Linguistic Data Consortium (LDC2017T01) \cite{kato-etal-2016-construction} that covers \textit{compound words} used to train parsing models, and Wiki50 by \cite{vincze-etal-2011-multiword} which consists of 50 Wikipedia articles (4,350 sentences) with the annotation of MWEs (compound, verb-particle constructions, idiom, Light verb construction, multi-word verbs) and named entities (PER, LOC, ORG and MISC).
Also related to this are  English MWEs from ``web reviews data" by \cite{schneider-etal-2014-comprehensive} that covers \textit{noun, verb} and \textit{preposition super-senses} and English verbal MWEs from \cite{mwe2018english} and \cite{kato2018mwe} that covers PARSEME shared task defined vMWE categories. However, all these works were performed in monolingual settings, independently by different language speakers without any bilingual alignment. These corpora are helpful for monolingual MWE research such as \textit{discovery} or \textit{identification}, however, it would be difficult to use these corpora for bilingual or multilingual research such as MT or cross-lingual information extraction.

The work most related to ours is the one by \cite{lvc2012hungarian}, who created an English-Hungarian parallel corpus with annotations for light verb constructions (LVCs). As many as 703 LVCs for Hungarian and 727 for English were annotated in this work, and a comparison between English and Hungarian data was carried out. 
LVCs often have a noun and a stretched verb, where the noun keeps its literal senses while the verb ``loses its original sense to some extent'' \cite{lvc2012hungarian}, for instance, ``take a bite out of...'' which is similar to the verb ``bite'', ``did cleaning'', and ``taken into consideration'', etc. However, the work did not cover other types of vMWEs, for instance inherently adpositional verbs, verbal idioms, or verb-particle constructions, and it was not extended to any other language pairs. 
In our work, we annotate in a multilingual setting including distant languages such as German (Germanic), Polish (Slavic), Italian (Romance), Chinese, and Arabic, in addition to the extension of vMWE categories. 
In other recent work \cite{Han2020multimwe}, we performed an automatic construction of bilingual MWE terms  based on a parallel corpus, in this case, English-Chinese and English-German. We first conducted automated extraction of monolingual MWEs based on part-of-speech (POS) patterns and then aligned the monolingual MWEs into bilingual
terms based on statistical lexical translation probability. However, due to the automated procedure, the extracted bilingual ``MWE terms" contain not only MWEs but also normal phrases. Part of the reason for this is due to the POS pattern design which is a challenging task for each language and needs to be further refined \cite{Skadina2016MultiwordEI}, \cite{rikters2017mwe}, \cite{Han2020multimwe}. 

In summary, in this work we aim at creating a multilingual parallel corpus that contains broader vMWE categories introduced by PARSEME shared task \cite{mwe2017sharedtask,PARSEME2018task,mwe-2020-joint}  but in a parallel and multilingual setting. This will benefit both the MWE identification task by extending the languages covered and down-stream multilingual applications especially MT since there is still a lack of such data on MWEs-aware test sets for MT system testing, especially high-quality and manually annotated ones \cite{lvc2012hungarian,Skadina2016MultiwordEI,Han2020multimwe}. 

The potential applications of AlphaMWE include its integration into MT systems similar to  \cite{tan-pal-2014-manawi}, who applied bilingual MWE and named entities corpora into English-Hindi (EN-HI) and Hindi-English (HI-EN) MT, and  \cite{zaninello-birch-2020-multiword,Han2020multimwe}, who performed MWE aware NMT.
Task-oriented MT evaluation is still in need of creating corresponding data sets \cite{jair2020MTE}, and AlphaMWE may be  used for such a purpose, namely to test an MT model's ability for translating idiomatic MWEs as one of the main indicators to reach human parity. 
For instance, the very recent work by \cite{Bawden2020jlre} created an English-French corpus for MT evaluation on a dialogue task; \cite{macketanz-etal-2021-linguistic} emphasised idioms as challenges for English$\leftrightarrow$German MT systems in WMT2021 shared task;
our own work on human-centric and task-oriented metric HOPE \cite{Gladkoff_Han_HOPE} using professional post-editing annotation for MT evaluation (MTE) uses optimised error categories which can be deployed using the AlphaMWE corpus for testing MT engines and we will give an example testing using English-Arabic corpus.
AlphaMWE would also be an asset for MWE identification and information extraction in a multilingual scenario, e.g. \cite{vincze-etal-2011-multiword} conducted  identification of noun compounds and named entities with the monolingual MWE corpus they prepared.


\begin{figure*}[t]
\begin{center}
\centering
\includegraphics*[width=\textwidth]{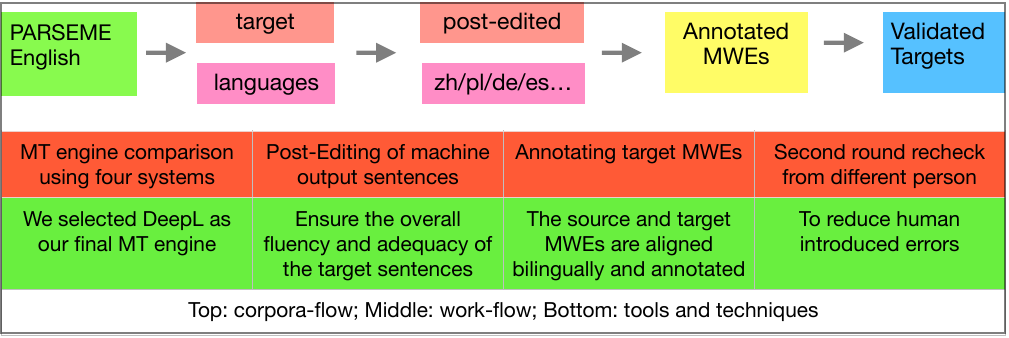}
\caption{Workflow to prepare the AlphaMWE corpus} 
\label{fig:AlphaMWE_workflow}
\end{center}
\vspace{-1ex}
\end{figure*}

\section{Corpus Construction: AlphaMWE}
\label{corpus_construction_section}

We now describe our corpus preparation method, the selection of the MT models used in our investigation, and the resulting open-source corpora AlphaMWE.

\subsection{\textit{Corpus Preparation}}

To construct a well-aligned multilingual parallel corpus, our approach is to take a monolingual corpus from the PARSEME \textit{vMWE discovery and identification shared task} as our root corpus. 
Our rationale  is
that this shared task is well established and its process of tagging and categorisation is clear. 
Furthermore, as we plan to extend the MWE categories in the future e.g., not only the verbal ones, we enrich the PARSEME shared task corpus with  potential for other downstream research and applications, including bilingual and multilingual NLP models such as MT. The English corpus \cite{mwe2018english} we used from the PARSEME shared task follows the annotation guidelines having a broad range of vMWE categories tagged. These include inherently adpositional verbs, light verb constructions, multi-verb constructions, verbal idioms, and verb-particle constructions. The English corpus contains sentences from several different domains such as news, literature, and IT documents, which were observed during our human post-editing procedure. 
For the IT document domain, vMWEs are usually easier or more straightforward to translate, with a high chance of repetition, e.g. ``apply filter" and ``based on", which are mostly translated correctly by the MT engines. 
For the literature annotations, the vMWEs include richer samples with many idiomatic or metaphorical expressions, such as ``cutting capers" and ``gone slightly to someone's head" that cause MT issues. 

Fig.~\ref{fig:AlphaMWE_workflow} shows our workflow. 
We first use MT models to perform automated translation for the target language direction and then carry out human post-editing of the output hypotheses with annotation of the corresponding target-side vMWEs which are aligned with the source English ones. 
In order to choose which MT system to use for producing the translations, we perform manual
inspection on 10 (randomly picked) outputs from these systems: DeepL, GoogleMT, Bing Translator, and Baidu Translator.
The rationale for the MT+PE design is that firstly the state-of-the-art NMT engines can produce relatively good results even though often with a certain degree of mistakes, which already saves a lot of time in comparison to human translation from scratch; secondly, one of the motivations of our corpus construction is to apply it to a multilingual NLP setting especially MT, and we are curious to examine how good (or bad) the MT performance is in term of MWEs.
Finally, to avoid human-introduced errors, we apply a cross-validation strategy, where each sentence receives at least a second person's quality checking after the first post-editing. 
Tagging errors are more likely to occur if only one human has seen each sentence and we discuss some error samples from English source corpus in later sections. 

The instructions given to the post-editors and annotators include i) edit any mistranslations that occurred with a meaning-equivalency aspect from source English to the target language; ii) wherever possible, try to use corresponding MWEs in the target language for the terms that are highlighted as MWEs in the source English; iii) offer alternative translations regarding some phrase/expression translations wherever suitable. iv) try to keep the register-level the same from the source language to the target language, e.g. `thx' \textit{vs} thanks, a word of lower register that occurs in some social media data; v) wherever an ambiguous translation occurs, the background context information shall be referred to, which is available to all post-editors.
Overall the post-editing and annotation tasks were evenly shared among workers except for English-Chinese 
and English-Italian where more experienced annotators have carried out larger amount of annotations than less experienced annotators.

The background of the post-editing and annotation workers are a) English-German: two research students who are German native speakers including one master's student and one Ph.D. candidate with backgrounds in linguistics and informatics (both graduated now); b) English-Polish: two Ph.D. candidates who are Polish native speakers with a background of linguistics and semantics (`semantics' one graduated); c) English-Chinese: four Chinese Mandarin native speakers including two Ph.D. candidates from applied linguistics and NLP (`NLP' one graduated), and two full-time faculty from the applied linguistics and child language learning areas (both hold a Ph.D. degree); d) English-Italian: two native Italian speakers one of whom is a postdoctoral researcher in computer science and informatics, and the other works in Ireland for 5 years in customer service sector as a fluent English speaker; e) English-Arabic: one Ph.D. candidate in NLP focusing on MWEs and one Masters student in Translation Studies.




\subsection{\textit{MT System Selection}}

\begin{figure*}[!t]
\begin{center}
\centering
\includegraphics*[width=\textwidth]{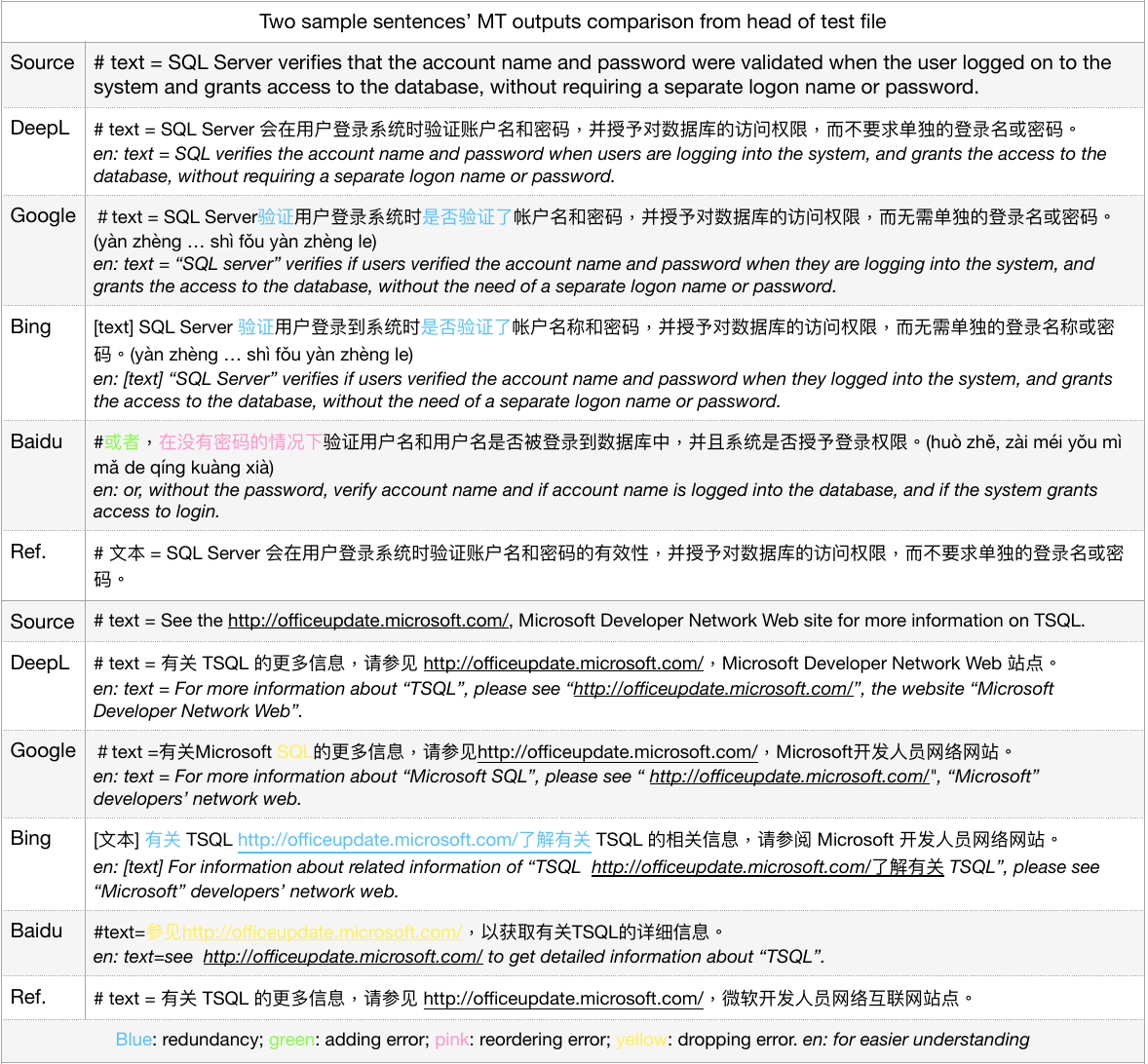}
\caption{Sample comparison of outputs from four MT models} 
\label{fig:compare_4mt}
\end{center}
\end{figure*}

We tested a number of example sentences (approximately 10) from the English test set to compare state-of-the-art MT from Microsoft Bing \cite{msr2017mt}, GoogleMT \cite{google2017attention}, Baidu Fanyi \cite{baidu2019mt}, and DeepL\footnote{https://www.deepl.com/en/translator (All testing was performed in 2020/07 from 4 MT models)}, as in Fig.~\ref{fig:compare_4mt}.\footnote{All testing was performed in 2020/07}. 
We illustrate the comparative performances with two representative example translations, which reflect the quality differences between these engines on the corpus we are using. As a first example sentence, GoogleMT and Bing Translator have very similar outputs, where the MT engines have a tendency to produce as much information as possible, but make the sentences redundant or awkward to read,
such as the phrase ``\textbf{验证}...是否\textbf{验证}了\,(yàn zhèng ... Shì fǒu yàn zhèng le)" where they use a repeated word ``验证" (yàn zhèng, \textit{verify}). Although the DeepL Translator does not produce a perfect translation since it drops 
the source word ``validated" which should
be translated as “有效性 (yǒu xiào xìng)” (as one candidate translation), the overall output is fluent and the source sentence meaning is mostly preserved. Baidu translator yields the worst output in this example. It produces some words that were not in the source sentence (或者, huò zhě, \textit{or}), loses some important terms’ translation from the source sentence (“SQL Server”, the subject of the sentence), and the 
reordering of the sentence fails resulting in an incorrect  meaning (``在没有密码的情况下, zài méi yǒu mì mǎ de qíng kuàng xià" is moved from the end of the sentence to the front and made as a condition). So, in this case, DeepL performed best.

In a second example sentence, GoogleMT confused the original term TSQL as SQL. Bing MT had a similar issue with the last example, i.e. it produced redundant information ``有关 (yǒu guān)" (\textit{about/on}).
In addition,  it concatenated the website address and normal phrase ``了解有关\, (liǎo jiě yǒu guān)" together with a hyperlink. GoogleMT and Bing both
translated half of the source term/MWE ``Microsoft Developer Network Web" as ``Microsoft开发人员网络网站" (kāi fā rén yuán wǎng luò wǎng zhàn) where they kept “Microsoft" but translated “Developer Network Web”. Although this is a reasonable output since Microsoft is a general popular named entity while ``Developer Network Web" consists of common words, we interpret ``Microsoft Developer Network Web" as a named entity/MWE in the source sentence that consists of all capitalised words which would be better translated overall as ``微软开发人员网络网站 (wēi ruǎn kāi fā rén yuán wǎng luò wǎng zhàn)" or be kept as the original capitalised words as a foreign term in the output, which is how
DeepL outputs this expression.
However, Baidu totally drops out this MWE translation and another word translation is not accurate, ``more" into 详细 (xiáng xì).
These samples illustrate why
we chose to use DeepL as the provider of our MT translations. Similar comparisons were carried out on English-German and English-Polish which also favoured DeepL. 

\subsection{\textit{Outcome: AlphaMWE Corpus}}

\begin{figure*}[!t]
\begin{center}
\centering
\includegraphics*[width=\textwidth]{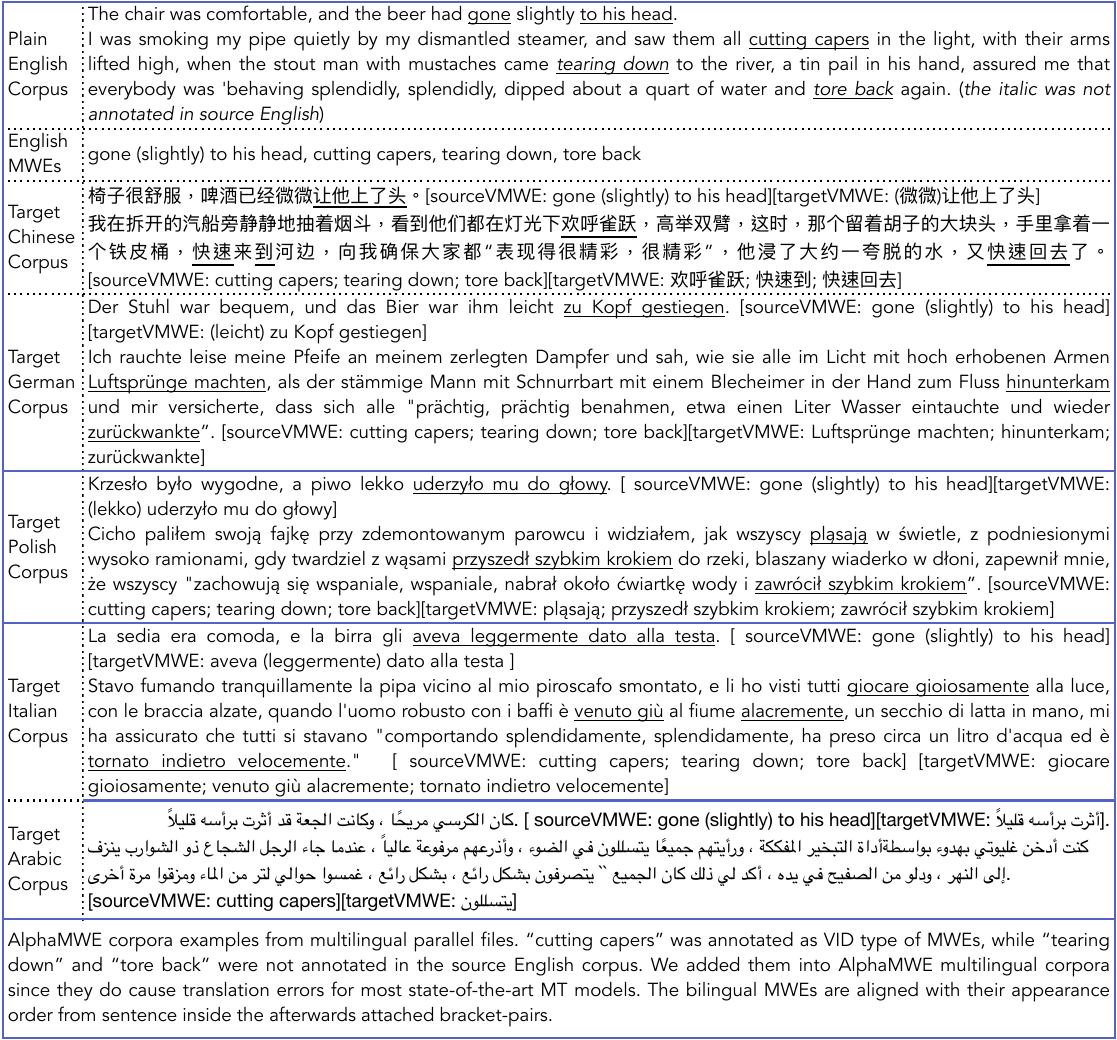}
\caption{AlphaMWE corpora samples with two sentences (English, Chinese, German, Polish, Italian, Arabic)} 
\label{fig:AlphaMWE_samples}
\end{center}
\end{figure*}

Regarding the size of the corpus, we extracted all 750 English sentences which have vMWE tags included. The target languages  are Arabic, Chinese, German, Polish, and Italian with sample sentences in  Fig.~\ref{fig:AlphaMWE_samples}. 
Our very recent work on confidence sample size estimation for MT evaluation using Monte Carlo sampling simulation indicates that the test set with 200+ sentences can reflect MT quality in a sufficient matter \cite{serge_et_al_2021measuring}.
This means that our bilingual/multilingual corpora with 750 sentences are suitable for shared task usage, specifically to a linguistic phenomenon.

There are several situations and decisions that are worth noting during the post-editing (PE) stage:
a) when the original English vMWEs are translated into a general phrase in the target language but not choosing a sequence of MWEs, we tried to offer two different references, with one of them being revised in a vMWE/MWE presentation in the target. For instance ``(the beer had) slightly gone to his head'' can be translated as ``微微让他上了头 \, (wéi wéi ràng tā shàng le tóu, he got a little drunk)" or ``他微微醉了 \, (tā wéi wéi zuì le, he got a little drunk)" of which the first term ``上了头 (shàng le tóu, literally `up to the head')" is an idiomatic MWE describing the concept ``醉了 \, (zuì le, literally ``drunk")"; b) when the original English sentence terms were correctly translated into the target language but in a different register, e.g. the source language has a low register (`thx', for instance), we asked the post-editors to offer two reference sentences, with one of them using the same low register and the other with (formal) full word spelling; 
c) for the situations where
a single English word or normal phrase is
translated into a typical vMWE in the target language, or 
both source and target sentences include vMWEs but the source vMWE was not annotated in the original English corpus, we made some additions to include such vMWE (pairs) into AlphaMWE; d) for some wrong/incorrect annotation in the source English corpus or some mis-spelling of words, we corrected them in AlphaMWE; e) we chose English as root/source corpus, since 
the post-editing and annotation of target languages require the human annotators
to be fluent/native in both-side languages, and all annotators
were fluent in English  as well as
being native speakers in the specific target languages respectively.

As shown in the examples (Fig.~\ref{fig:AlphaMWE_samples}) from Chinese, German, Polish, and Italian, 
all languages are sentence-aligned and the paired vMWEs (source, target) are listed at the end of each sentence in the order in which they appear in the sentence.
AlphaMWE also includes statistics of the annotated vMWEs and a multilingual vMWEs glossary. The AlphaMWE corpora were divided evenly into five portions which were designed in the post-editing and annotation stage. The sentences were grouped in the order that they were extracted from the original English source larger corpus with context information. 
As a result, it is convenient for researchers to use them for testing NLP models, choosing any subset portion or combination, or cross-validation usage.



\section{MT issues with MWEs}
\label{sec_MT_issues}
As mentioned in the previous section, one of the motivations for creating the AlphaMWE multilingual corpus with MWEs is to apply it to multilingual NLP applications such as MT and examine how MT performs in an MWE-related context. 
In this section, we present an analysis of the quality of various MT systems when required to translate MWEs or MWE-related contexts. We performed the categorisation of errors during the early post-editing stage, and this categorisation was presented to all the post-editing and annotation workers during frequent communications  across the editing phase.

In this paper, we mainly focus on the English$\rightarrow$Chinese language pair. We also highlight
some issues on English$\rightarrow$German, English$\rightarrow$Polish, and English$\rightarrow$Arabic in different sections, but
leave a detailed analysis of other language pairs for 
future work. Some extra examples of English-Polish MT issues related to MWEs will be presented in the Appendix section. 

\subsection{\textit{English-to-Chinese}}
When MT produces incorrect or awkward translations this can fall into   different categories, and from our analysis, we classify  them as \textit{common sense, super sense, abstract phrase, idiom, metaphor, and ambiguity}, with ambiguity further sub-divided into context-unaware ambiguity, social/literature-unaware ambiguity, and coherence-unaware ambiguity. These classifications are to be further refined in the future, e.g., the differences between metaphor and idiom are sometimes fuzzy.
We now list each of these with examples to support
future MT research 
on improving the quality of MT when handling MWEs. 



\subsubsection{Common Sense}

\begin{figure*}[!t]
\begin{center}
\centering
\includegraphics*[width=\textwidth]{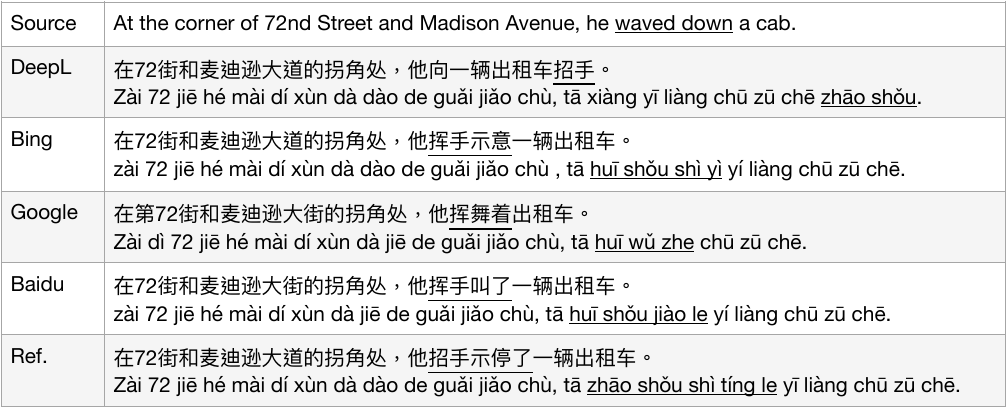}
\caption{MT issues with MWEs: common sense (Pinyin is offered by GoogleMT with post-editing.)} 
\label{fig:mwe_figure_common_sense}
\end{center}
\end{figure*}

The first error category is called  \textit{common sense}. This category of error occurs when the translation models did not acquire any common knowledge or social context that human beings will take for granted, and that can be used to carry out reasoning wherever suitable.
For instance, the sentence in Fig.~\ref{fig:mwe_figure_common_sense} includes the vMWE ``waved down" which in general understanding indicates that ``he succeeded in getting the cab" and not only ``waved his hand". However, in the translation by DeepL and Bing this vMWE was wrongly translated as ``he waved his hand to the cab" missing part of the original meaning; the MT output by GoogleMT is also incorrect, saying ``he waves with the cab in hand"; the Baidu translation of this sentence is semantically correct that ``he waved and got one cab" though it does not use a corresponding Chinese side vMWE ``
招手示停(zhāo shǒu shì tíng)"\footnote{We give full sentence pronunciation (in Pinyin) of Chinese characters in this figure, for the following examples, we only present the Chinese Pinyin for MWEs and studied words of the sentences to save space.}.


\subsubsection{Super Sense}

\begin{figure*}[!t]
\begin{center}
\centering
\includegraphics*[width=\textwidth]{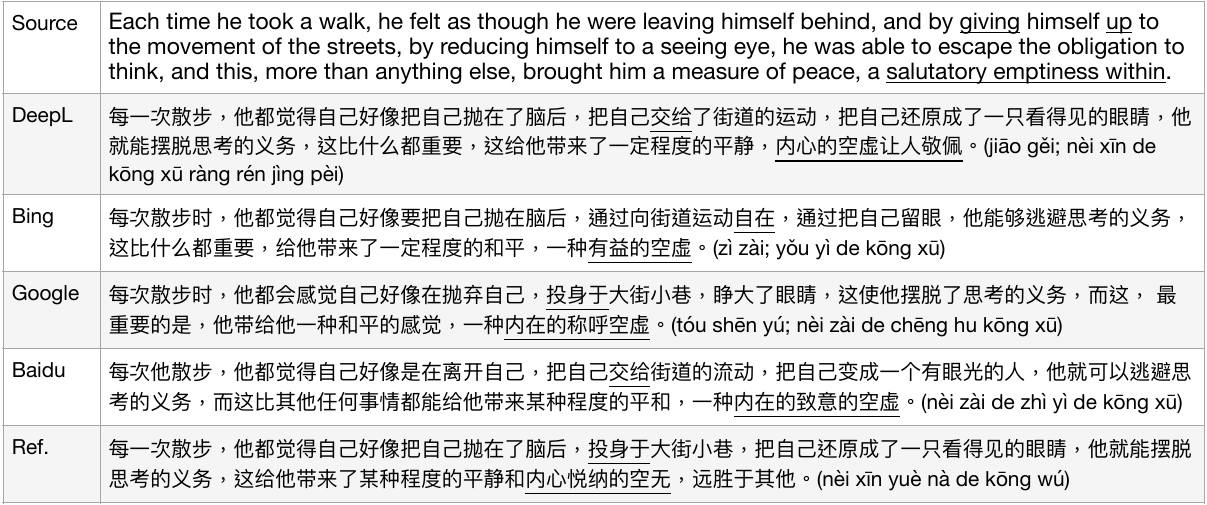}
\caption{MT issues with MWEs: super sense } 
\label{fig:mwe_figure_super_sense}
\end{center}
\end{figure*}

This category of translation issue is related to \textit{a form of state of mind} and we need to make a logical prediction to guess the correct interpretation, e.g. \textit{positiveness}, \textit{negativeness}, \textit{neutral}, or other situations for some words, in the choice of Chinese characters.
As in Fig.~\ref{fig:mwe_figure_super_sense}, 
the MT systems each have  advantages for different parts of this long sentence. However, none of them is perfect. For instance, for the translation of vMWE ``giving (himself) up (to)", the DeepL and Baidu outputs give very literal 
translation by saying ``he gives himself to", 
the Bing translator drops
the vMWE, while GoogleMT preserves the 
correct meaning in the 
translation ``投身于\,(tóu shēn yú)" from 
the reference indicating ``he devoted himself". However, GoogleMT’s output for the phrase ``salutatory emptiness within" is very poor and makes no sense; the reference is ``the emptiness that he welcomes" for which Baidu has a closer translation ``内在的致意的空虚\,(nèi zài de zhì yì de kōng xū)". All four MT outputs also use the same Chinese words ``空虚\,(kōng xū)" which is a term with a negative meaning, however, the sentence indicates that he is welcoming this emptiness, which should be the corresponding Chinese word ``空无\,(kōng wú)", 
an unbiased or positive meaning.

\subsubsection{Abstract Phrases}

\begin{figure*}[!t]
\begin{center}
\centering
\includegraphics*[width=\textwidth]{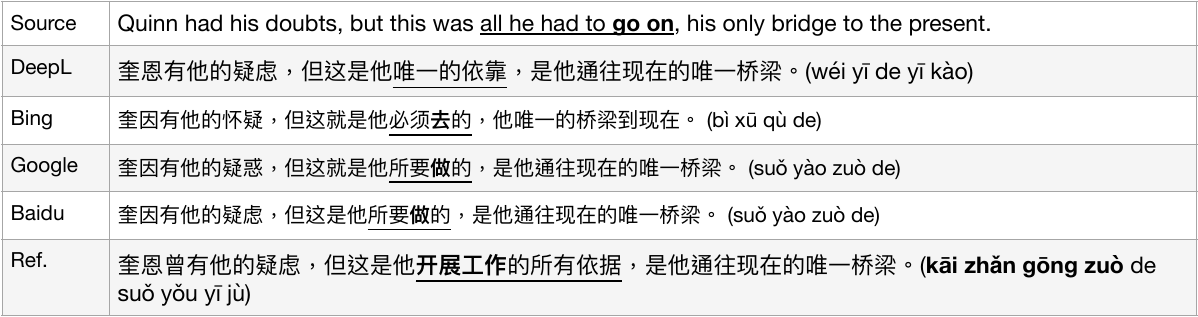}
\caption{MT issues with MWEs: abstract phrases} 
\label{fig:mwe_figure_abstract_phrase}
\end{center}
\vspace{-2ex}
\end{figure*}

The \textit{abstract phrases} can have different exact meanings and we usually need some background information from the sentence or paragraph to select the correct word choices in the target language\footnote{it sometimes belongs to the context-unaware ambiguity (CUA) that we will mention later, however, CUA not necessarily means ``abstract phrase", and usually needs paragraph information, not only sentence level. Furthermore, in some situations, we just don't know how to interpret ``abstract phrase", i.e. the candidate interpretations are unknown without context, and this is different from ambiguity.}. 
As mentioned in Section \ref{corpus_construction_section}, in our post-editing and annotation task, the background context information of extracted sentences was given to the workers.
With the example sentence in Fig.~\ref{fig:mwe_figure_abstract_phrase},
from the context, we know that ``go on" in this sentence means ``to work from" using all the information he had. The phrase ``this was all he had to go on" is then to be interpreted as ``this is all the information he had to work from". At the end of the sentence, ``the present"  is the ``present person" he needs to look for (with the picture of this person’s younger age portrait). However, Bing translated it as ``this is (where) he had to go" which is an incorrect  interpretation
of ``had to go"; furthermore, Bing's translation of the second half of the sentence kept the English order, without any reordering between the words, which is grammatically incorrect in Chinese, i.e. ``他唯一的桥梁到现在 (tā wéi yī de qiáo liáng dào xiàn zài)". GoogleMT and Baidu translated it as ``what he need to do" which is also far from correct, while DeepL successfully translated the part ``his only thing to relying on" but dropped 
the phrase ``go on", i.e., \textit{to do what}. \textit{Abstract Phrase} can include \textit{Super Sense} as its sub-category, however, it does not necessarily relate to a state of mind. 


\subsubsection{Idioms}

\begin{figure*}[!t]
\begin{center}
\includegraphics*[width=\textwidth]{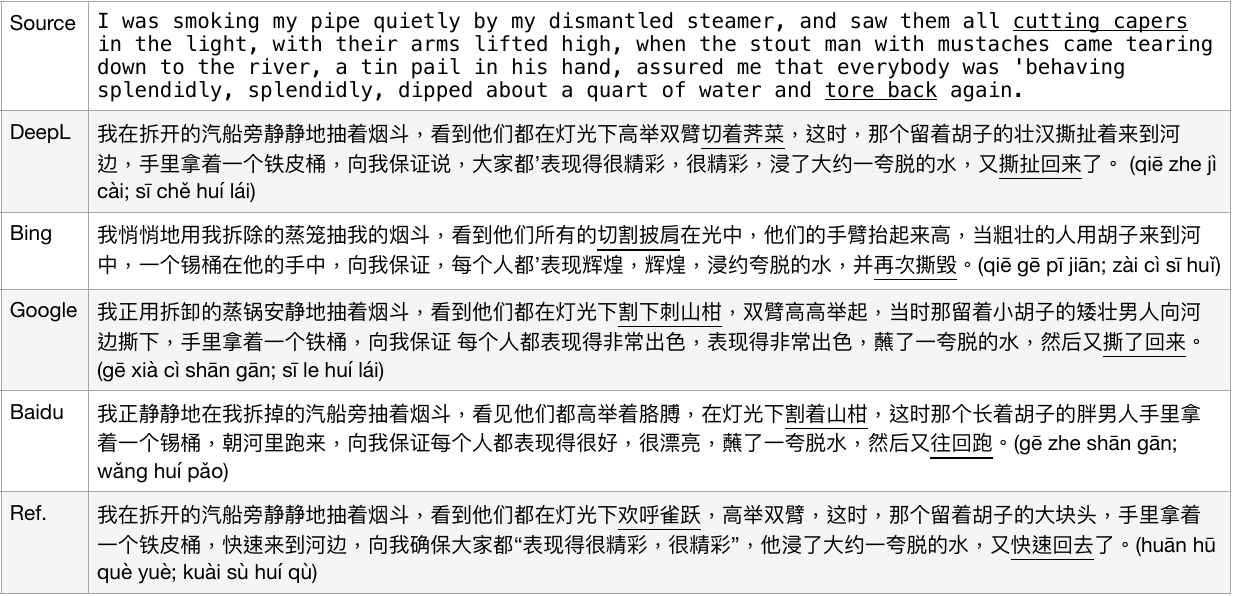}
\caption{MT issues with MWEs: idioms 
\label{fig:mwe_figure_idiom}}
\end{center}
\end{figure*}

\begin{figure*}[!t]
\begin{center}
\includegraphics*[width=\textwidth]{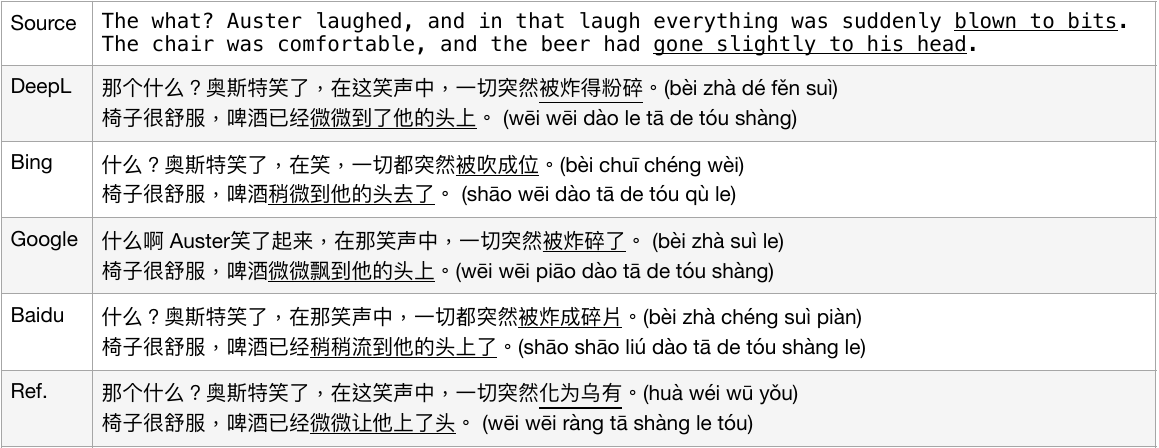}
\caption{MT issues with MWEs: metaphor 
\label{fig:mwe_figure_metaphor}}
\end{center}
\end{figure*}

The use of \textit{idioms}  often  causes  wrongly translated sentences, mostly resulting in humorous output due to literal translation. For example, in the sentence in Fig.~\ref{fig:mwe_figure_idiom}, the vMWEs ``\textit{cutting capers}" and ``tore back" are never translated correctly at the same time by any of the four  MT models we used. The idiom ``cutting capers" indicates frolic or romp, to ``act in the manner of a young goat clumsily frolicking about" \footnote{https://www.dictionary.com/browse/cut-capers}, 
and here it means ``they are in a happy mood, playful and lively movement" which should properly  be translated as the corresponding Chinese idiom ``欢呼雀跃 (huān hū què yuè, happily jumping around like sparrows)". However, all four MT models translated it literally into ``cutting" actions just with different objects, i.e., what they cut. The idiom (slang) ``tore back" means the stout man \textit{walked back rapidly}, which the Baidu translation gives the closest translation as ``往回跑 (wǎng huí pǎo, run back)" but the other three models translated into an action ``tear something (to be broken)" which is incorrect.

\subsubsection{Metaphors}

The first sentence vMWE ``blown to bits" in Fig.~\ref{fig:mwe_figure_metaphor} is a \textit{metaphor} to indicate ``everything is gone", instead of the physical ``blowing action". However, the three MT models DeepL, GoogleMT, and Baidu translate it as
``exploded into pieces (by bombs)", while
BingMT translates it even more literally into ``blown to (computer) bits". There is a corresponding Chinese vMWE ``化为乌有(huà wéi wū yǒu, vanish into nothing)" which would be a  proper choice for this source vMWE translation. 

The second sentence vMWE ``gone (slightly) to his head" is a metaphor to indicate ``got slightly drunk". However, all  four MT models translate it as physically ``beer moved to his head" but by slightly different means such as \textit{flow} or \textit{flutter}. The corresponding translation as a MWE should be ``微微让他\textbf{上了头} (wéi wéi ràng tā \textbf{shàng le tóu})", using the same characters, but the character order here makes so much difference, meaning ``slightly drunk".

\subsubsection{Ambiguity}
We summarised different kinds of situations that cause ambiguity in the resulting translation when it meets MWEs or named entities, so we further divide ambiguity into three sub-classes.


\textbf{I).Context-Unaware Ambiguity}

\begin{figure*}[!t]
\begin{center}
\centering
\includegraphics*[width=\textwidth]{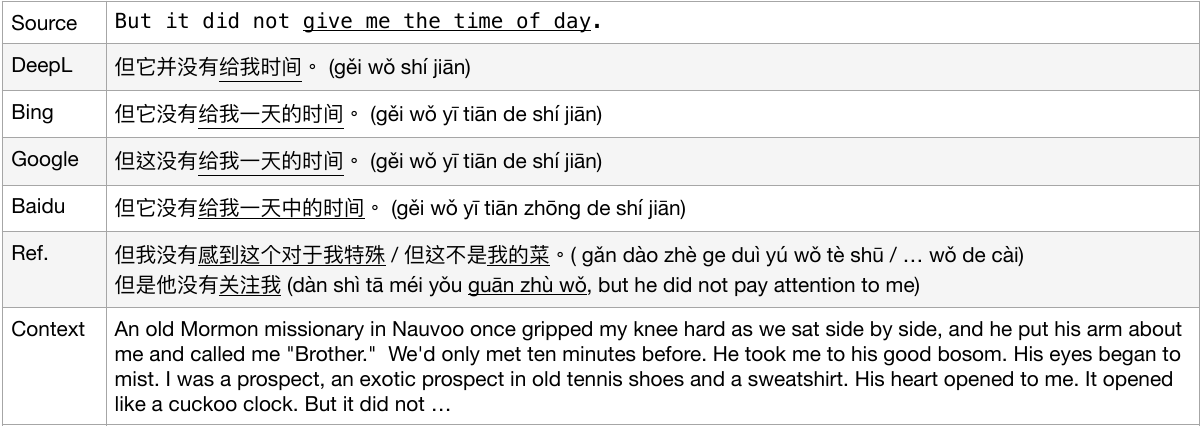}
\caption{MT issues with MWEs: context-unaware ambiguity} 
\label{fig:mwe_figure_ambiguity_content}
\end{center}
\end{figure*}

In this case, the \textit{context}, i.e. the background information, is needed for the correct translation of the sentence. For instance, see Fig.~\ref{fig:mwe_figure_ambiguity_content}. DeepL gives the translation ``it did not give me time though", while Bing and GoogleMT give the same translation ``it/this did not give me one day’s time" and Baidu outputs a grammatically incorrect sentence. From the pre-context, we understand
that it means the speaker ``did not feel that is special to him" or ``did not have an affection of that" after \textit{all the Mormon missionary’s effort towards him}. Interestingly, there is a popular Chinese idiom (slang) that matches this meaning very well ``不是我的菜 (bù shì wǒ de cài, literally \textit{not my dish})". We also offer an alternative translation in our corpus for this sentence as ``但是他没有关注我 \, (dàn shì tā méi yǒu guān zhù wǒ, but he did not pay attention to me)" since there is this idiom ``not to give someone the time of day" indicating ``not to pay attention to someone" and it also makes sense in this story context.
From this point of view, the context-based MT model deserves some more attention, instead of only focusing on sentence level. When we tried to put all background context information as shown in Fig.~\ref{fig:mwe_figure_ambiguity_content} into the four MT models, they produce the same output for this studied sentence, as for sentence level MT. 
This indicates that current  MT models still focus on sentence-by-sentence translation when meeting paragraphs, instead of using context inference;
or alternatively, that even if the system is able to utilise multi-sentence information, that information is not sufficient to resolve the correct translation at least in this case.

\begin{figure*}[!t]
\begin{center}
\centering
\includegraphics*[width=\textwidth]{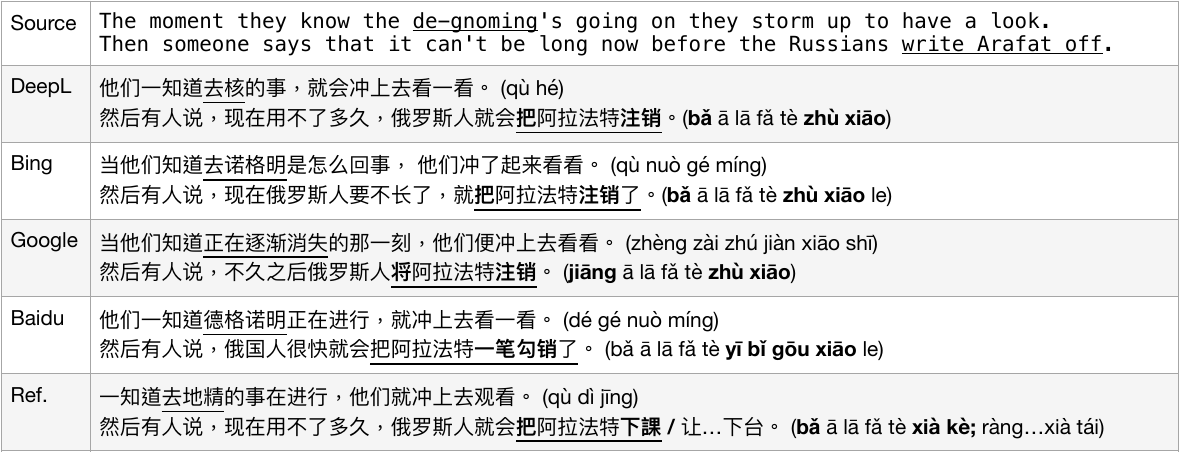}
\caption{MT issues with MWEs: social/literature-unaware ambiguity} 
\label{fig:mwe_figure_ambiguity_social_literature}
\end{center}
\end{figure*}

\textbf{II).Social/Literature-Unaware Ambiguity:\\}
In this case,  \textit{social knowledge} of current affairs from news, or  \textit{literature knowledge} about some newly invented entities and phrases are required in order to get a correct translation output. For instance, Fig.~\ref{fig:mwe_figure_ambiguity_social_literature} includes two sentences, one from politics and another from literature.

In the first sentence, ``de-gnoming" is a literature word from Harry Potter, invented by its  author, to refer to the process of ridding a garden of gnomes, \textit{a small magical beast}. 
Without this literature knowledge, it is not possible to  translate the sentence correctly. For instance, even though this sentence is from a very popular novel that has been translated into most languages, DeepL translated it as
``去核 (qù hé, de-nuclear)", Bing translated it as ``去诺格明 (qù nuò gé míng, \textit{de}-nuògémíng" where ``nuògémíng" is a simulation of the pronunciation of ``gnoming" in a Chinese way, Baidu translated it as
``德格诺明 (dé gé nuò míng)" which is the simulation of the pronunciation of the overall term ``de-gnoming".
As we mentioned in an earlier section on corpus preparation, we add examples of translation pairs where we think it is a challenge to MT, including words with corresponding translations as MWEs in the target language. 
In this sentence, we treat the target Chinese translation “去地精 (qù dì jīng)” as an MWE with a special structural construction “去 (qù, \textit{get rid of}) + noun” indicating “remove/delete + noun”. Since the word “de-gnoming” is a literature word, the Chinese MWE “去地精  (qù dì jīng)” is also regarded as a borrowed foreign word.
The original vMWEs annotated in this English sentence are “going on” (VPC.full) and “have (a) look” (LVC.full); however, in the MT task, apparently these two vMWEs have been addressed better by MT engines than the literature word “de-gnoming”, even though the Bing translator also made a mistake on translating "going on" into "是怎么回事 (shì zěn me huí shì, what is going on)".

In the second sentence, ``write Arafat off" is to dismiss ``Yasser Arafat", Chairman of the Palestine Liberation Organization, which is a historical person’s name. However, all  three models DeepL, Bing, and GoogleMT translated it into ``把/将阿拉法特注销 (bǎ/jiāng ā lā fǎ tè zhù xiāo, \textit{deregister Arafat})" which treated ``Arafat" as a title of certain policy/proceeding, not being able to recognize it as a personal named entity, while Baidu made the effort to use the Chinese idiom ``一笔勾销 (yī bǐ gōu xiāo, \textit{cancel everything}, or \textit{never mention historical conflicts})" for ``write off" but it is not a correct translation. Interestingly, if we put these two sentences into a web search engine  it retrieves the  correct web pages as the context in the top list of the search result. 
This may indicate that  future MT models could consider including web search results as part of their knowledge of background for translation purposes, e.g. using generative translation models \cite{Radford2019LanguageMA_GPT2,T5}. 

\textbf{III).Coherence-unaware ambiguity}
\begin{figure*}[!t]
\begin{center}
\centering
\includegraphics*[width=\textwidth]{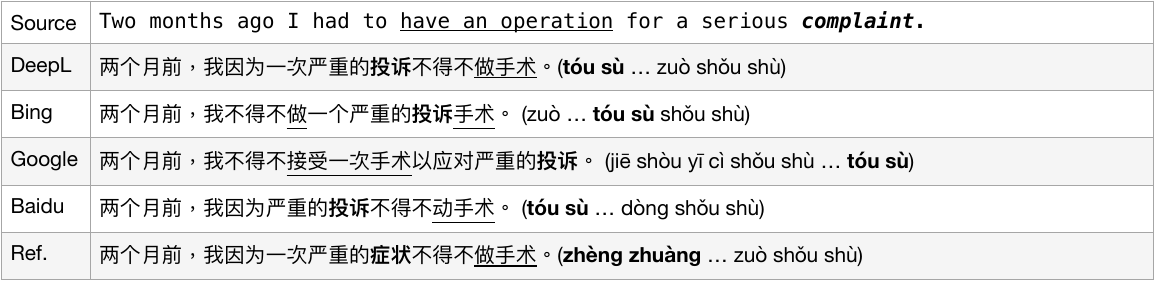}
\caption{MT issues with MWEs: coherence-unaware ambiguity} 
\label{fig:mwe_figure_ambiguity_coherence}
\end{center}
\end{figure*}

This kind of MWE ambiguity can be solved by the \textit{coherence} of the sentence itself, for instance, the example in Fig.~\ref{fig:mwe_figure_ambiguity_coherence}. The four MT models all translated the vMWE itself ``have an operation" correctly in meaning preservation by ``做/接受/动 手术 (zuò/jiē shòu/dòng shǒu shù)" just with different Chinese word choices. 
However, none of the MT models translated the ``reason of the operation", i.e., ``complaint" correctly. 
The word complaint has two most commonly used meanings ``a statement that something is unsatisfactory or unacceptable" or ``an illness or medical condition" and all four models chose the first one. 
According to the simple logic of social life, people do not need to ``have an operation" due to ``a statement", instead their ``medical condition" should have been chosen to translate the word ``complaint". 
Because of the incorrectly chosen candidate translation of the word ``complaint", Bing’s output even invented a new term in Chinese ``投诉手术 (tóu sù shǒu shù, \textit{a surgery of complaint statement kind})" which makes no sense.
This raises the issue of the ``cognitive plausibility'' of current powerful language models in NLP fields.

\subsection{\textit{English-to-German}}

In the case of English-to-German MWEs, there are some cases where the corresponding German translation of English MWEs can be one word. This is partial because  German has compound 
verbs. For instance, the vMWE ``woke up" the sentence ``An old woman with crinkly grey hair woke up at her post outside the lavatory and opened the door, smiling and grasping a filthy cleaning rag." has corresponding German aligned word ``erwachte" with a suitable translation “Eine alte Frau mit krausem, grauem Haar erwachte auf ihrem Posten vor der Toilette und öffnete die Tür, lächelte und griff nach einem schmutzigen Putzlappen.”.


This ``MWE to single-word'' alignment phenomenon might not be specific for the English-German language pair, since this also occurs in the English-to-Chinese and English-to-Arabic translation, such as an English verb+particle MWE being aligned to one single Chinese character/word.
For example, in this sentence ``The fact that my name has been mixed up in this.", the vMWE (VPC) \textit{mixed up} gets aligned to a single character word ``混\,(hùn)" in a suitable translation ``事实上，我的名字已经被混在这里面了。(shì shí shàng, wǒ de míng zì yǐ jīng bèi hùn zài zhè lǐ miàn le)".
However, in the view that German has many compound words and it might be more frequent than other languages, we list this issue under the English-German sub-category. 

A second issue is that the automatic translation to German can be  very \textit{biased} towards choosing the formal (or polite) form \textit{vs} informal form. See the examples such as “Sie” instead of the second form singular “du” for ``you", ``auf Basis von" instead of ``basierend auf" for ``based on". 
To achieve a higher level of translation accuracy, MT models need to choose the correct or more suitable form of words depending on the context being used.

A third issue is that,
for the English verbal multi-word expressions that are often not translated as verbal multi-word expressions to German, this indicates some further work to explore by MT researchers to develop better models to have the machine producing corresponding German existing MWEs.

\subsection{\textit{English-to-Polish}}
The MT output issues in English-to-Polish mostly fall into the categories of coherence-unaware error, literal translation, context-unaware error, and gender-related mistakes, in addition to other errors similar to what we listed in the Chinese-English category.

Regarding the MT output issues on English-to-Polish that fall into coherence-unaware error, for instance, the vMWE ``write off" in the sentence ``Then someone says that it can't be long now before the Russians write Arafat off." was translated as ``Wypiszą" (Potem ktoś mówi, że już niedługo Rosjanie wypiszą Arafata.) which means ``prescribe", instead of the correct  ``spiszą na straty (Arafata)". This error can be avoided by the coherence of the sentence itself in meaning preservation models.

For the literal translation, we can see the example vMWE ``gave (him) a look" in the sentence ``She ruffled her feathers and gave him a look of deep disgust." which was literally translated as ``dała mu spojrzenie", however, in Polish, people use ``\textbf{throw} a look" as ``rzuciła (mu) spojrzenie" instead of ``gave (dała, a female form)"\footnote{a proper translation: \textit{Nastroszyła sobie pióra i rzuciła mu spojrzenie głębokiego obrzydzenia.} Also the MT output word for ``Nastroszyła" was ``Zdruzgotała" which has the wrong meaning. 
}. 
Another example of literal translation leading to errors is the vMWE ``turn the tables" from the sentence ``Now Iran wants to turn the tables and is inviting cartoonists to do their best by depicting the Holocaust." which is translated as ``odwrócić stoliki (turn tables)", however, it should be ``odwrócić sytuację (turn the situation)" or ``odwrócić rolę (turn role)" with a proper translation ``\textit{Teraz Iran chce odwrócić sytuację i zachęca rysowników, by zrobili wszystko, co w ich mocy, przedstawiając Holocaust}." 
These two examples illustrate the localisation issue in the target language.


For the context unaware issue, we can look back to the example sentence ``But it did not give me the time of day." from Fig.~\ref{fig:mwe_figure_ambiguity_content}. This was literally translated word by word into ``Ale nie dało mi to pory dnia." which is in the sense of hour/time. However, it should be ``Nie sądzę aby to było coś wyjątkowo/szczególnie dla mnie. (I do not think this is special to me.)" based on the context, or ``Ale to nie moja bajka" as an idiomatic expression which means ``not my fairy tale" (indicating \textit{not my cup of tea}).

Regarding gender-related translation errors, for instance,
in some translations, the word ``I'' in English shall be reconsidered if it specifies a male or female which results in different translations in the target Polish. 

We listed some examples of mistranslation of gender-related terms in Fig. \ref{fig:gender_error_pl} (Section Appendix
). The word “friend” in English can be both male and female, but in Polish male is “przyjaciel” and the female “przyjaciółka”. The example sentence “I have a friend” was translated using the male form “mam przyjaciela” instead of the female form “Mam przyjaciółkę” even though there is the context of “her” indicating the friend is a female gender. 
Also, the English word “that” in this situation is gender specific in Polish as well, because it leads to a clause relating to “friend”,  and “przyjaciel” and “przyjaciółka” are conjugated (because of the verb “have” followed).





\section{Quantifying MT Errors using English-Arabic for Case Study}
\label{sec_en-ar-HOPE}
\begin{figure*}[!t]
\begin{center}
\centering
\includegraphics*[width=\textwidth]{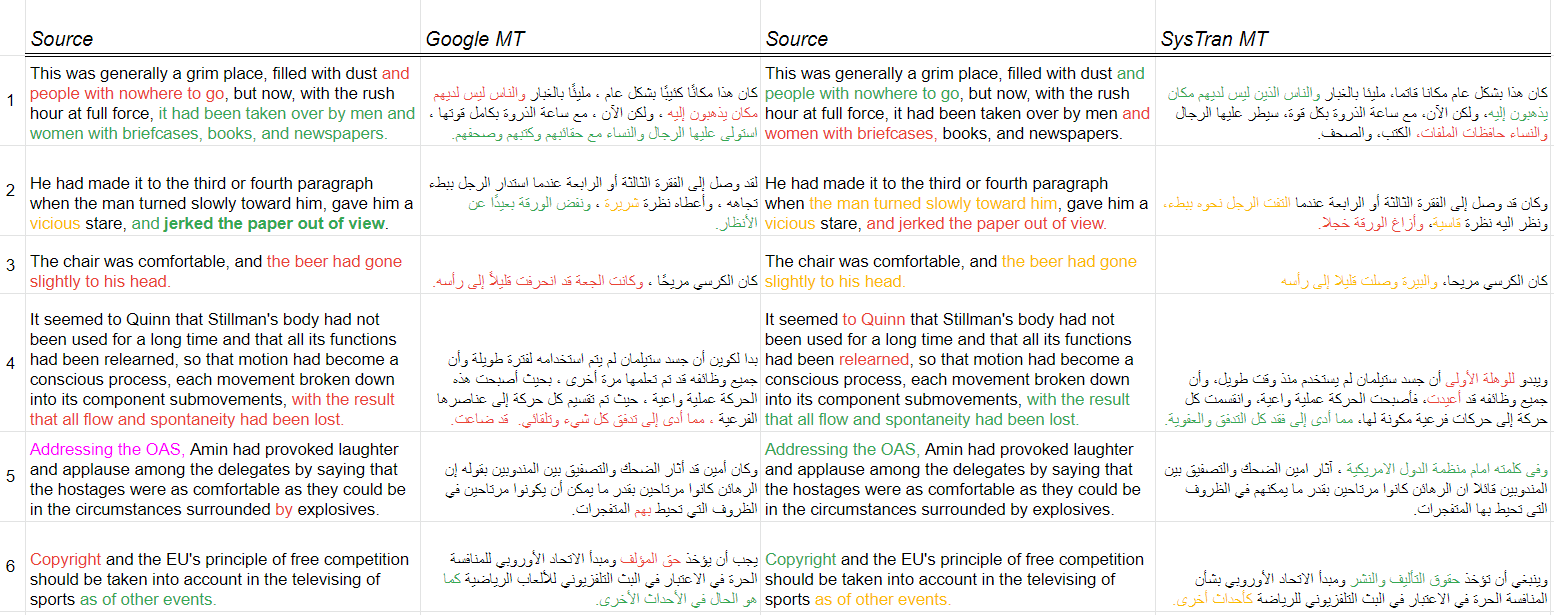}
\caption{MT Output Comparisons between GoogleMT and Systran (Green: well translated, Red: wrong translation, Yellow: correct but unnatural and Magenta: skipped.)} 
\label{fig:google-vs-systran-MT}
\end{center}
\end{figure*}

For Arabic corpus creation, firstly there is no Arabic language option in the DeepL Translator. Following the same procedure on the corpus creation of other languages, by comparing MT systems and selecting one as raw output provider for next step post-editing, 
our two native Arabic speakers performed some manual comparison on GoogleMT versus SysTran MT \footnote{\url{https://www.systransoft.com/lp/machine-translation/}} and we show some working examples in Figure \ref{fig:google-vs-systran-MT}. 
In these examples, we find that GoogleMT produced more \textit{meaning}-correct outputs in comparison to the SysTran engine, even though GoogleMT made more mistakes on \textit{entity} translations. 
We thought entity errors are easier to correct and this procedure also gives us the chance to look into MWE-related errors by MT engines.
So we picked GoogleMT for Arabic corpus creation.

\subsection{\textit{Revisiting HOPE Metric}}

To qualitatively and quantitatively evaluate the MT output of English-to-Arabic translation, i.e. how many errors the state-of-the-art MT system makes in this language pair on our data and to what severity levels, we deployed the HOPE metric \cite{Gladkoff_Han_HOPE} for this purpose. 
There are eight original error types designed in HOPE, which include Mistranslation (MIS), Style (STL), Terminology (TRM), Impact (IMP), Missing Required Adaptation, Ungrammatical (UGR), Proofreading Error (PRF), and Proper Name. 
To reflect our task more precisely, we added two more error types during our post-editing and scoring, which are ``MWE Missed Chance (\textbf{MMC})'' and ``Skipped Word (\textbf{SKP})''. 
The MMC error is to quantify the situation when the source-side MWEs are translated either wrongly in the target, or correctly in meaning but the model missed the chance of choosing the correct target MWEs when there are indeed such MWEs in the standard Arabic language.
The SKP error is to highlight the situation when GoogleMT skipped some important source words during the translation procedure.

The HOPE metric has a hierarchical scoring procedure, from the segment level to the system level. Each error type is also assigned with specific penalty scores according to the severity level, using $2^n$ (1, 2, 4, 8, 16) for ``minor, medium, major, severe, and critical'' levels.
The segment-level score is the sum of each individual error score. 
To classify the errors in a more straightforward view, the segment-level scores are grouped into three categories, either minor (score 1-4), major (score 5+), or correct.

There are a few key indicators in HOPE metric.
1) The total penalty score (T.P.S.) for each type of error is the sum of all penalty scores overall segments. 
2) The ratio of total segments (R.o.T.) is the percentage of specific error types out of all errors T.P.S. 
3) The PPS value means the penalty point per segment that is the value of all penalty scores T.P.S. 
divided by segment numbers, 
i.e. $PPS=Sum_{penalty}/Sum_{segment}$.

\subsection{\textit{HOPE Scoring Output}}

The HOPE error score marking was performed simultaneously when the post-editing was carried out for each segment.
To reflect details on different error types, we list the total penalty scores (T.P.S.) and their ratio (R.o.T.) out of all segments in Table \ref{tab:errors} using 150 segments. The T.P.S. value is 455 and the number of segment is 150.
From this detailed error type analysis, we can see that the MMC and SKP error types that we added to the HOPE metric occupied 17 percent and 6 percent of errors.
This also reflected that MWE-related errors take nearly 20 percent of all errors in this testing output.
This finding implicitly implies the importance of our multilingual corpus creation with MWE annotations that can be used for MT model validations.

\begin{figure*}
        \centering
         \includegraphics[scale=0.40]{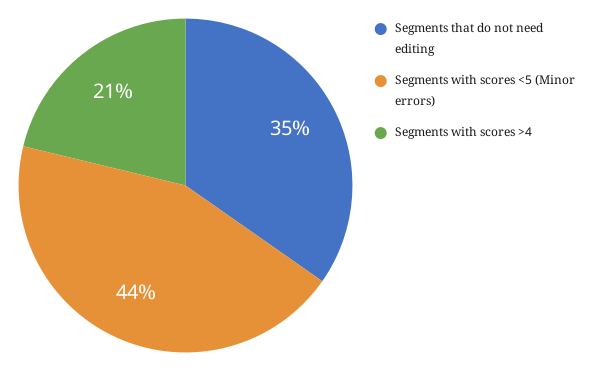}
         \caption{Quantitative Evaluation on GoogleMT Outputs from English to Standard Arabic using HOPE Metric: Minor Errors \textit{vs} Major Errors 
         }
          \label{fig:seg_percent}
         \end{figure*}

The statistics of GoogleMT output on error categories are shown in Figure \ref{fig:seg_percent}, which tells that 21 percent of the output segments fall into major errors that need to be fixed, 44 percent of segments fall into minor errors that can be good enough for some applications, and only 35 percent of them are correct translations that do not need to edit.

\begin{table*}[!t]
    \label{hope_metric}
    \centering
    \begin{tabular}{ccccccccccc}
    \hline
    \textbf{Types} &MMC& MIS &STL & TRM & IMP& UGR & PRF & SKP & All & PPS \\
    T.P.S. &76 &	68&	69&	39	&114	&37	&46	&6	&455 & \\
    R.o.T. &17\% &15\% &15\% &9\% &25\% &8\% &10\% &1\% && 3.03
\\    \hline
\end{tabular}
\caption{Detailed Error Types and Ratios (T.P.S.: total penalty scores. R.o.T.: Ratio out of total segments.)} \label{tab:errors}
\end{table*}

\section{Discussion}
\label{sec_Discussion}

In this section, we present some broader issues that we discovered during our corpus construction process, which are related to MWEs and MT.

\subsection{\textit{Issues in the source English corpus}}

Some problems occurred in the source English corpus which impact the sentences we extracted from the monolingual training and test data with vMWE annotation tags.

Firstly, there is an accuracy issue with the tagging, which may have been introduced as a result of the tagging task being carried out with one person per sentence. Each sentence, either long or short, is located in one line. \footnote{https://gitlab.com/parseme/sharedtask-data/-/tree/master/1.1/EN}.  

Some error annotations of vMWEs in the source monolingual corpus will thus have some impact on the accuracy level of the \textit{vMWE discovery and identification} shared task but also affect the bilingual usage of AlphaMWE, so we tried to address all these cases. For instance, in the example sentence in Fig.~\ref{fig:mwe_figure_abstract_phrase},  the English corpus annotated wrongly the sequence ``had to go on" as a verbal idiom (VIDs) which is not accurate. The verb ``had" here is affiliated with ``all he had" instead of ``to go on". So either we  annotate ``go on" as vMWE in the sentence or the overall clause ``\textit{all he had to go on}" as a studied term. In AlphaMWE, we reserved “(to) go on” as the vMWE studying term, since “all he had to go on” is more like a sentence clause than a vMWE term.

Another example with a different type of vMWE is the sentence ``He put them on in a kind of trance." where the source English corpus tagged ``put" and ``trance" as Light-verb construction (LVC.cause). 
However, the phrase is with ``put...on" instead of ``put...trance". 
The phrase ``put someone into a trance" is  to express ``make someone into a half-conscious state". However, for this sentence, if we check back a bit further of the context, it means ``he put on his cloth in a kind of trance". The word ``trance" is affiliated with the phrase ``\textit{in a kind of trance}" instead of ``put".


%

\begin{figure*}[!t]
\begin{center}
\centering
\includegraphics*[width=\textwidth]{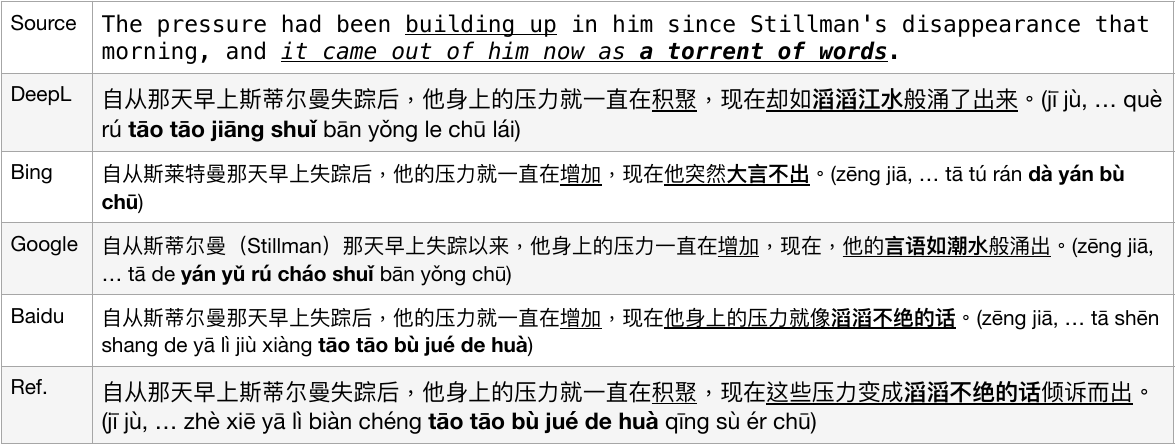}
\caption{Additional vMWEs or MWEs} 
\label{fig:mwe_figure_additional}
\end{center}
\end{figure*}

A second issue we discovered with the English language corpus is that there are some interesting sentences in the  corpus that include non-decomposable MWEs but these MWEs are not annotated. We plan to add further annotation on this aspect and extend this kind of bilingual pairs.

For instance, in Fig.~\ref{fig:mwe_figure_additional}, the vMWE category verb-particle constructions (VPC.semi) is tagged to the phrase ``building up", which may be an interesting case for vMWE discovery and identification, however, for cross-lingual research, as the initial aim of our corpus construction, such as MT, the ending part of this sentence ``came out of him now as a torrent of words" poses more challenges, and would draw more attention from researchers. 

We test this with four MT models and found the following outputs: DeepL literally translated it into ``a strong and fast-moving stream of water” and dropped ``words”; Bing gave a translation with the opposite meaning ``does not say a word”; Google and Baidu produced much better translation covering ``torrent of words” even though the sentence level translation contain errors and can be improved, i.e., ``it” meaning the ``pressure” in the source sentence was dropped out by Google; ``came out of him” was dropped out by Baidu.


\subsection{\textit{Broader findings on MT issues with MWEs}}

In this section, we introduce some broader findings on MT issues and interesting phenomena related to MWEs that still exist for the current state-of-the-art models, focusing on the English-to-Chinese pair.

\subsubsection{Entity Translation Issue}

\begin{figure*}[!t]
\begin{center}
\centering
\includegraphics*[width=\textwidth]{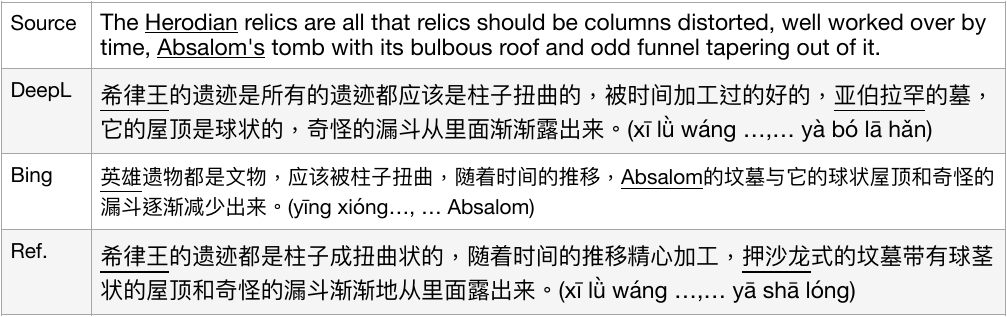}
\caption{General MT: named entity} 
\label{fig:mwe_figure_entity}
\end{center}
\end{figure*}

Named entities (NEs) from  Western languages always match into  multi-character / word expressions in the Chinese language, even though it is a single word in the original Western form, e.g. the family name or surname 
of a person. 
In some documentary stories with some famous names, including ones from Bible, the mistranslation of named entities will lead to inaccurate history, and possibly cause arguments. For instance, from the Bible, \textit{Absalom} was the son of David, the King of Israel, while \textit{Abraham} was the founding father of the ``Covenant of the pieces", the special relationship between the Hebrews and God \footnote{https://www.thebiblejourney.org}. In this example sentence in Fig.~\ref{fig:mwe_figure_entity}, the named entity ``Herodian” was correctly translated by DeepL as ``希律王 (xī lǜ wáng)”, however, wrongly translated as ``hero” by Bing only taking part of this word ``Hero" instead of ``Herod" from Herod the Great. The named entity ``Absalom” was kept as an unknown word by Bing without translation, and was wrongly translated by DeepL into ``亚伯拉罕 (yà bó lā hǎn)”, which is actually from another named entity ``Abraham”, and the correct translation of it in Chinese is the multi-character/word sequence ``押沙龙 (yā shā lóng)” \footnote{https://biblehub.com http://biblehub.net/searchchin.php?q=押沙龙}. 

\subsubsection{The English-Style Chinese Patterns}
The English-style Chinese situation occurred when the MT models face some translation of English phrase patterns.
We name this as \textit{English-style Chinese} or \textit{EngliChinese} in short, which means the Chinese sentence sounds like the English pattern-based expression, such as either the syntax of the Chinese sentence is in English, or there are some English input words (or literally translated words) that apparently make the sentence sound awkward in Chinese.

\begin{figure*}[!t]
\begin{center}
\centering
\includegraphics*[width=\textwidth]{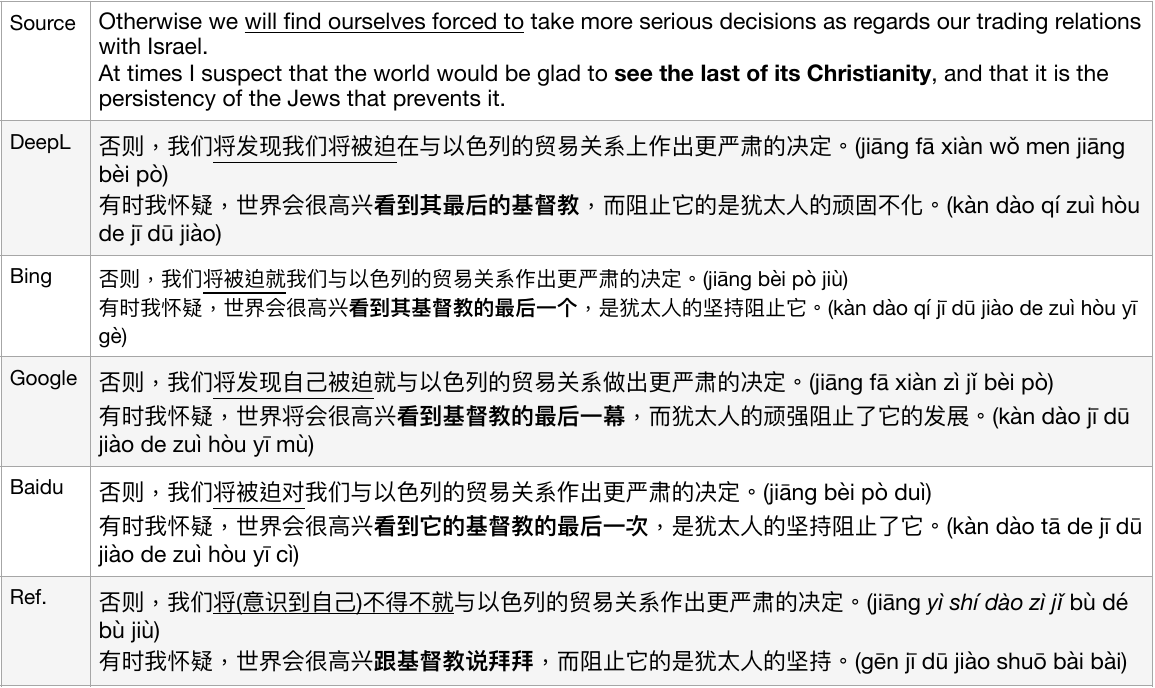}
\caption{General MT: 
\textit{English-style} Chinese} 
\label{fig:mwe_figure_EngliChinese}
\end{center}
\end{figure*}

For the first example sentence, Fig.~\ref{fig:mwe_figure_EngliChinese} ``will find ourselves (forced to) …” is an English pattern that shall not be translated just word by word into Chinese. However, DeepL and Google both made the same mistake by the literal translation into Chinese, either dropping ``find ourselves” in the translation or replacing ``find” as ``realize” which would make more sense in Chinese, such as ``将(意识到自己)不得不, jiāng (yì shí dào zì jǐ) bù dé bù“.
The second sentence, ``see the last of its Christianity” contains one idiom ``see the last of (someone)”, however, DeepL, Bing, and Baidu all translated it into an awkward Chinese sequence, especially Baidu, which literally translated it word by word from English  plus a literal moving of ``its” to the front.


\subsubsection{Mis-use of MWEs}
This issue is also related to MWEs. The current SOTA MT models  are apparently able to incorporate some target side MWEs into their candidate output. However, some MWEs on the target side were wrongly picked up by the MT engine which resulted in the translation having a different meaning to the source.

\begin{figure*}[!t]
\begin{center}
\centering
\includegraphics*[width=\textwidth]{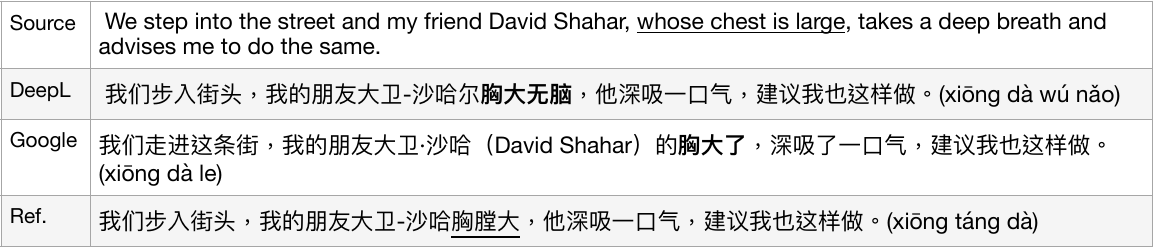}
\caption{General MT: misusage of MWEs} 
\label{fig:mwe_figure_misUseMWE}
\end{center}
\end{figure*}

The output of DeepL with the example sentence in Fig.~\ref{fig:mwe_figure_misUseMWE} used a  Chinese idiom ``胸大无脑 (xiōng dà wú nǎo, literally ``big chest, no brain”) to translate ``whose chest is large” which is  wrong by virtue of adding some extra meaning that does not exist in the source text. This may be caused by the training corpora that DeepL used which have a higher probability to align ``chest ... large ...” to ``胸大无脑\,(xiōng dà wú nǎo)”.
In addition, the Google output is not correct either by using ``胸大了 (xiōng dà le, chest becomes large)” which is different from the original meaning.

\subsection{\textit{Document-level MT for MWEs}}

\begin{figure*}[!t]
\begin{center}
\centering
\includegraphics*[width=\textwidth]{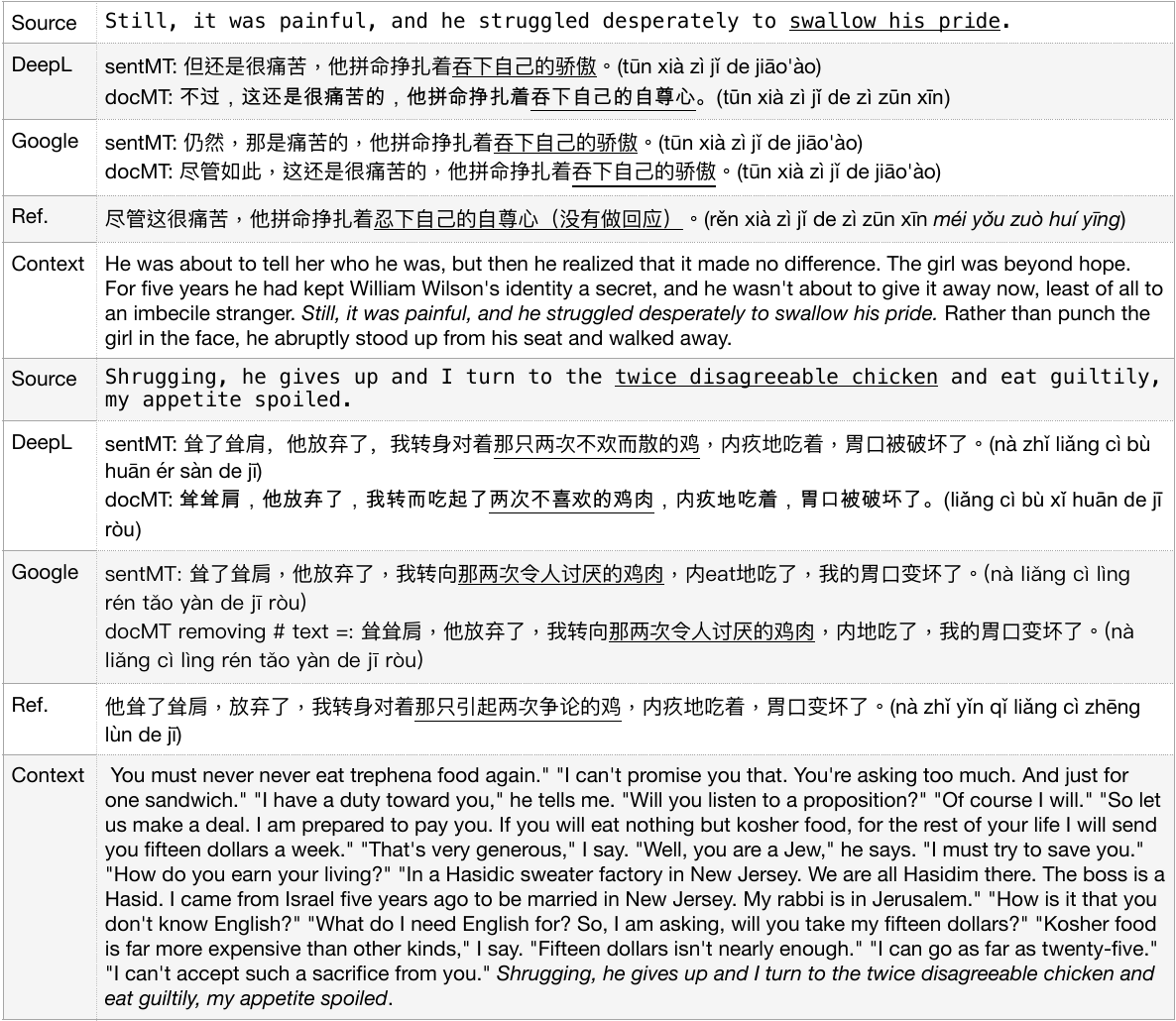}
\caption{General MT: document-level MT (docMT) vs sentence-level MT (sentMT)}
\label{fig:mwe_figure_docMT}
\end{center}
\end{figure*}

In the context-unaware ambiguity section, we mentioned document level MT as having some issues and here we give  further analysis on that topic. 
There are some phrases and MWEs that usually need context information to understand correctly first, before translation. 
In some cases, it is ambiguity, while in others it is just out of the blue that we do not have a clue how to translate the terms or phrases being used, i.e. they are totally unexpected. 
We present two examples here, one vMWE as an idiom and another as a noun phrase (adj+adj+noun), 
in Fig.~\ref{fig:mwe_figure_docMT}. The examples indicate that the current SOTA MT models can't handle this well yet.
%

For the first sentence, the idiom ``swallow someone’s pride” means ``to decide to do something although it will make him/her feel embarrassed or ashamed", and in this context, it means he ``did not say anything”, instead, just walked away
\footnote{https://dictionary.cambridge.org/dictionary/english/swallow-your-pride}. Here we can translate it as ``忍下自己的自尊心 (rěn xià zìjǐ de zì zūn xīn)” or ``没有做回应 (méi yǒu zuò huí yīng)”. DeepL document level MT gave a good pick up using the word ``自尊心 (zì zūn xīn)”, however, ``吞下自己的自尊心 (tūn xià zìjǐ de zì zūn xīn)” is not a correct way to say this in Chinese. All other three MT models, including sentence level DeepL, and sentence/document level GoogleMT yielded the same literal translation ``吞下自己的骄傲 (tūn xià zìjǐ de jiāo ào)” which makes no sense in Chinese \footnote{We used sentence level translation outputs of these MT engines since our corpus contains 750 independent sentences that are extracted from the original larger context. However, to have a view of document-level MT performance, we tested some example sentences with context using document-level MT which option is available on the interface of such engines. }.

For the second sentence, from the context, we understand the point that ``the twice disagreeable chicken” means that ``the chicken caused twice disagreements”. One happened in the beginning ``You must never...I can't promise you that" and another happened in the end ``So, I am asking, will you take my ...I can't accept such a sacrifice from you."
However, the document-level MT of DeepL and Google both failed to translate this meaning correctly, even though they did make some difference compared with the sentence-level model.


\subsection{\textit{MT on Chinese Pinyin}}
Out of the four MT models we used, two of them offer the Chinese pronunciation in Roman alphabets, i.e., Google and Bing, and this is termed ``拼音\,(Pīn yīn)". The character ``拼\,(pin)" indicates writing, and ``音\,(yin)" indicates sounds/reading.
We discuss the issues when MT models offer Pinyin for Chinese characters, which are sometimes related to MWEs.

\begin{figure*}[!t]
\begin{center}
\centering
\includegraphics*[width=\textwidth]{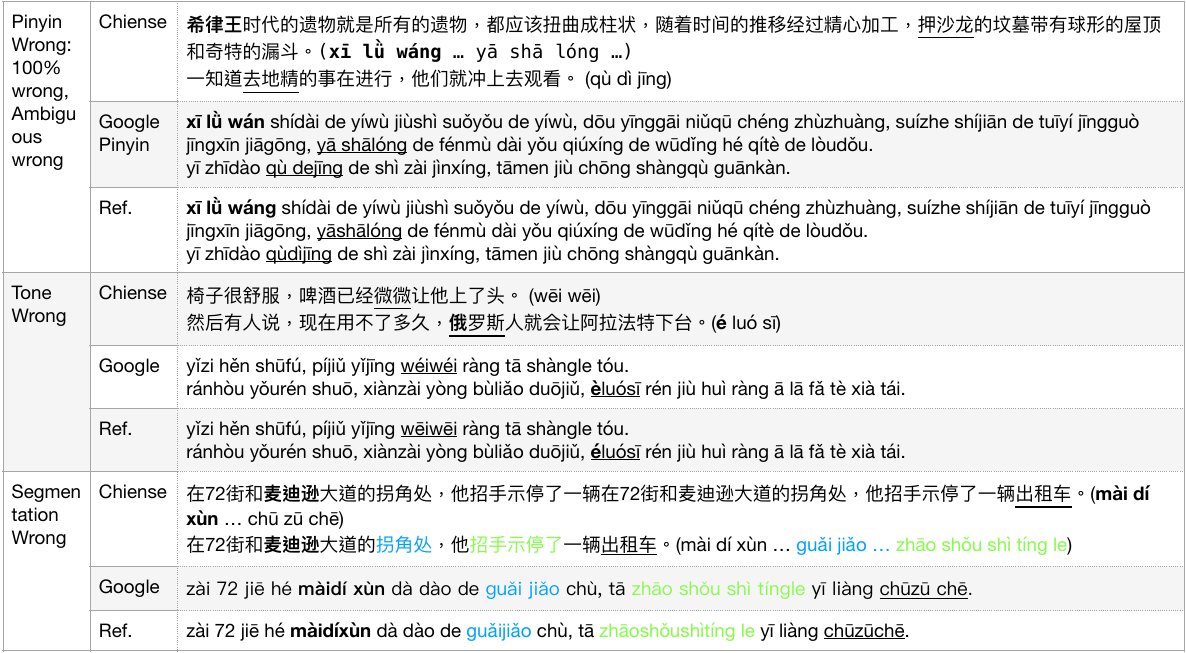}
\caption{General MT: Pinyin on MWEs}
\label{fig:mwe_figure_pinyin}
\end{center}
\end{figure*}

The Bing Translator just kept each Chinese character's Pinyin separated one by one, without any morphological segmentation. GoogleMT's output of Pinyin tried to perform  Chinese word segmentation, i.e., concatenate several characters' Pinyin together if they belong to the same word. 
However, there are several apparent errors in GoogleMT' Pinyin.
\begin{enumerate}
    \item  the Pinyin is wrong itself (totally wrong, or wrong choice in disambiguation situation);
\item the tone is wrong;
\item  the word segmentation is wrong which misleads the understanding of the sentence.
\end{enumerate}
In Fig.~\ref{fig:mwe_figure_pinyin}, we listed these three types of error with examples. The first error type was presented in two different situations, with one due to a wrong choice facing ambiguous pronunciation, and the other being totally wrong. 

The characters ``希律王\,(Xī lǜ \textbf{wáng}, Herodian)" in the sample sentence with the named entity issue  was wrongly annotated as ``Xī lǜ \textbf{wán}" by GoogleMT, which belongs to a totally wrong situation. \footnote{Furthermore, the entity ``押沙龙\,(yā shā lóng, Absalom)" was wrongly segmented into two pieces ``押\,(yā, bet)" and ``沙龙\,(shālóng, salon)". This leads to mis-understanding.} The second sentence of this error, ``去地精\,(qù dì jīng, \textit{de-gnoming})" was wrongly annotated as ``qù dejīng" which is due to the character ``地\,(dì)" which has different pronunciations\footnote{https://baike.baidu.com/item/地/34380 and https://en.wiktionary.org/wiki/地}. Due to the model failing to acquire the meaning of this term, it chose an incorrect pronunciation.




As another example, the reference sentence from the ``common sense" issue in Fig.~\ref{fig:mwe_figure_pinyin}, the name entity ``麦迪逊\,(mài dí xùn, Madison)" was wrongly segmented as ``麦迪\,(màidí)" and ``逊\,(xùn)". These two examples also reflect that the current MT models have issue in recognizing or processing \textit{foreign named entities}. 
In addition, the vMWE ``招⼿⽰停了\,(zhāo shǒu shì tíng le, waved down)" was incorrectly segmented as ``招|⼿|⽰|停了\,(zhāo shǒu shì tíngle)" which should be ``招⼿⽰停|了\,(zhāoshǒushìtíng le)". This illustrates the issue in accuracy level of automatic recognition of vMWE.
Furthermore, ``拐⾓\,(guǎi jiǎo, corner)" and ``出租车\,(chū zū chē, taxi)" should be grouped together as one concept respectively.


\subsection{\textit{Simplified vs Traditional Chinese MT}}

Traditional Chinese characters have rich linguistic knowledge and are naturally evolved from pictographs via thousands of years of civilization, while simplified Chinese characters are from recent history with less than one hundred years of usage in mainland China.  We think it is valuable to offer both these corpora, especially the traditional Chinese characters which are still being used by Taiwan, Hong Kong, Macau, and other regions, such as Kanji characters (made up of traditional Chinese) in Japan. 

\begin{figure*}[!t]
\begin{center}
\centering
\includegraphics*[width=\textwidth]{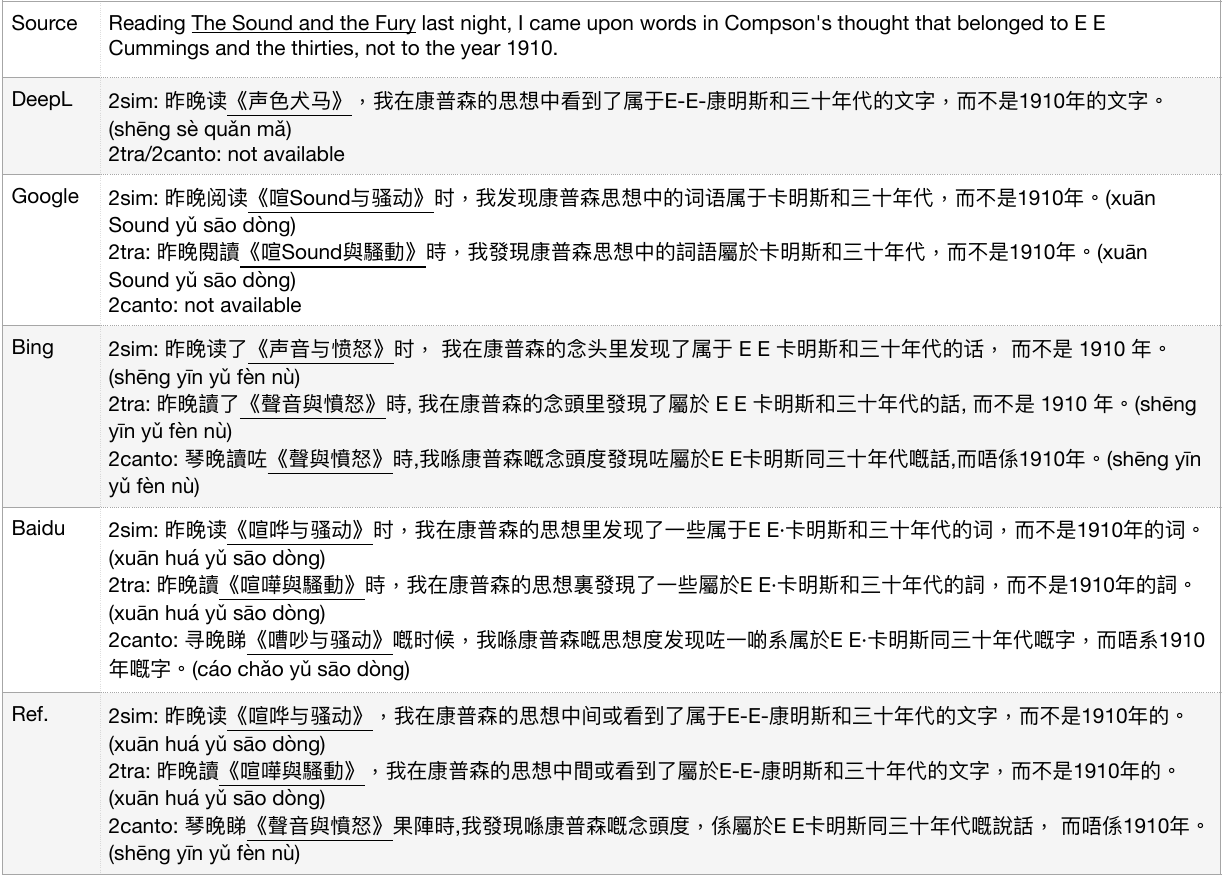}
\caption{General MT: Mandarin vs Cantonese, Simplified vs Traditional Chinese (2sim/tra/canto: English to simplified/traditional (Mandarin), and to Cantonese) }
\label{fig:mwe_figure_traditional_canto}
\end{center}
\end{figure*}

However, there are  some issues in the mismatch translation, for instance, some named entities are translated in different ways by simplified or traditional Chinese by mainland China, Hong Kong or Taiwan with examples in  Fig.~\ref{fig:mwe_figure_traditional_canto}.

In addition to writing system variations, there are different dialects in Chinese, including Mandarin from the Beijing area which is the official language, and Cantonese which is very popular and widely spoken in the Southern part of China including Guangdong province, Hong Kong, and Macau. Cantonese is originally from Guangdong province whose capital city called ``Guangzhou" was also known as ``Canton" in history.

Out of the four MT models, Bing and Baidu offer both traditional Chinese and Cantonese translation outputs. The difference is that Bing offers Cantonese in traditional Chinese characters, while Baidu offers it in simplified characters.
In the example sentence in Fig.~\ref{fig:mwe_figure_traditional_canto}, regarding the translation of <The Sound and the Fury>,
there are at least two issues with the Bing Translator at this point: firstly, when it translates into simplified Chinese, it actually uses the named entities from HK/TW translation <声音与愤怒, shēng yīn yǔ fèn nù>, just the simplified character of <聲音與憤怒, shēng yīn yǔ fèn nù>, which is a mess-up. Secondly, when it translates into Cantonese in traditional Chinese, it does not translate the term <聲音與憤怒, shēng yīn yǔ fèn nù> correctly by dropping <音, yīn>, in addition to that the overall sentence needs to be improved. 
Baidu’s translation of the Cantonese name <嘈吵与骚动, cáo chǎo yǔ sāo dòng> is incorrectly replacing ``聲音 \,(shēng yīn, sound)" with ``嘈吵 \, (cáo chǎo, noise)" and ``憤怒 \, (fèn nù)" with ``骚动 \, (sāo dòng)'' which is partly from the Mandarin translation. 

Regarding, the translation from DeepL, <声色犬马, shēng sè quǎn mǎ> is a 1974 Chinese movie from HK, which has nothing to do with the title <The Sound and the Fury> in the source sentence.

This literature translation issue reflects the difference between mainland China and other Chinese-speaking regions (Hong Kong, Macau, and Taiwan), also simplified vs. traditional Chinese characters. We need to make the corresponding changes and clarification when we prepare a traditional Chinese corpus for AlphaMWE.



\section{Conclusions and Future Work}
\label{sec_conclude}


In this paper, we presented the construction of a multilingual parallel corpus, AlphaMWE, with vMWEs as pioneer annotations by native speakers of the corresponding languages. 
We plan to extend our annotation into broader MWE categories that are non-verb ones, as well as some terms that might not count as MWEs but present challenges to MT systems, some of which were listed in the MT issues and Discussion section (e.g. ``de-gnoming"). 
We described the procedure of MT system selection, human post-editing, and annotation, compared different state-of-the-art MT models, and classified the MT errors from
vMWEs-related sentence/context translations. 
We characterised translation errors into different categories to help  MT research to focus on one or more of them to improve the performance of MT. For instance, how to integrate bilingual terminologies or dictionaries including paraphrases and synonyms to improve metaphorical and idiomatic phrase translation; or is it possible to design novel neural network structures to incorporate the MWE compositionality as part of the MT learning stage?

We performed the same process as described here for English$\rightarrow$Chinese, 
English$\rightarrow$German, 
English$\rightarrow$Polish,
English$\rightarrow$Italian, and 
English$\rightarrow$Arabic and similarly categorised the MT issues when handling MWEs.
We used English $\rightarrow$ Arabic corpus as a case study to quantify the error types and ratios using human expert-based post-editing metric HOPE.
We also included a future discussion session on corpus development, translation issues on named entities, Chinese Pinyin, Simplified vs Traditional Chinese, etc. where they are related to the processing of MWEs. 
In future work, we will conduct a more detailed analysis of the English-Italian corpus, in addition to English-German/Polish/Arabic. 

We name our corpus AlphaMWE to indicate that we will continue to maintain the developed corpora which are publicly available and extend them into other possible language pairs, e.g. the currently involved Spanish and French under-development. 
We also plan to extend the annotated MWE genres beyond the
vMWEs defined in the 
PARSEME shared task. 
There are some limitations on our corpus creation. For instance, even though we have carried out strict quality control on post-editing phase spacially on idiomatic MWEs translation, it might still bias the evaluation towards the MT system that we used to produce the initial raw candidate sentences when this corpus is used for MT system testing and evaluation. 

Our AlphaMWE multilingual corpora are available as  open access at \url{https://github.com/aaronlifenghan/AlphaMWE}.







\section*{Author Contributions}
LF carried out the initial work on multilingual resource creation of MWEs and co-coordinated the corpus annotation work including the annotation guidelines and recruiting native speakers for annotation. NHM and MR carried out Arabic corpus annotation and quantitative and qualitative evaluation using the HOPE metric. GJ and AS co-supervised the early work on AlphaMWE during LH's PhD and revised the initial manuscript. GN co-supervised the follow-up work on Arabic-MWE \cite{mohamed2023alphamwe_Arabic} and approved this extended manuscript.

\section{acknowledgements}
We thank the anonymous reviewers for their valuable comments and suggestions on improving this article. 
We are very grateful to each target language's post-editing and annotation contributors --- ZH: Pan Pan, Qinyuan Li, Ning Jiang; PL: Sonia Ramotowska, Tereska Flera; DE: Daniela Gierschek, Vanessa Smolik, Gültekin Cakir; IT: Gabriella Guagliardo, Paolo Bolzoni. 
We are thankful for the helpful discussion on English MWEs with Lorin, Roise, and Eoin, and on Arabic with Haifa. 
The extra examples on English-Polish MT with MWEs in the Appendix are mostly from SR. 
This work was partially funded by The ADAPT Centre for Digital Content Technology under the SFI Research Centres Programme (Grant 13/RC/2106).
The input of Alan Smeaton is partially funded by SFI under grant number SFI/12/RC/2289 (Insight Centre).
LH and GN were part-funded by grant EP/V047949/1 "Integrating hospital outpatient letters into the healthcare data space" (funder: UKRI/EPSRC).
LH is currently funded by the 4D Picture Project (\url{https://4dpicture.eu/}).

\section{Bibliographical References}\label{sec:reference}

\bibliographystyle{lrec2026-natbib}
\bibliography{lrec2026-example}

\appendix 

\section*{Extra examples on EN-PL MT issues including MWE-related}
\label{appendix}
We list more interesting examples from English-to-Polish translation, including social/literature-unaware ambiguity, super sense error, context-unaware error, metaphor translation error, abstract phrase MT error, and gender related MT errors.


\begin{figure*}[!h]
\begin{center}
\centering
\includegraphics*[width=\textwidth]{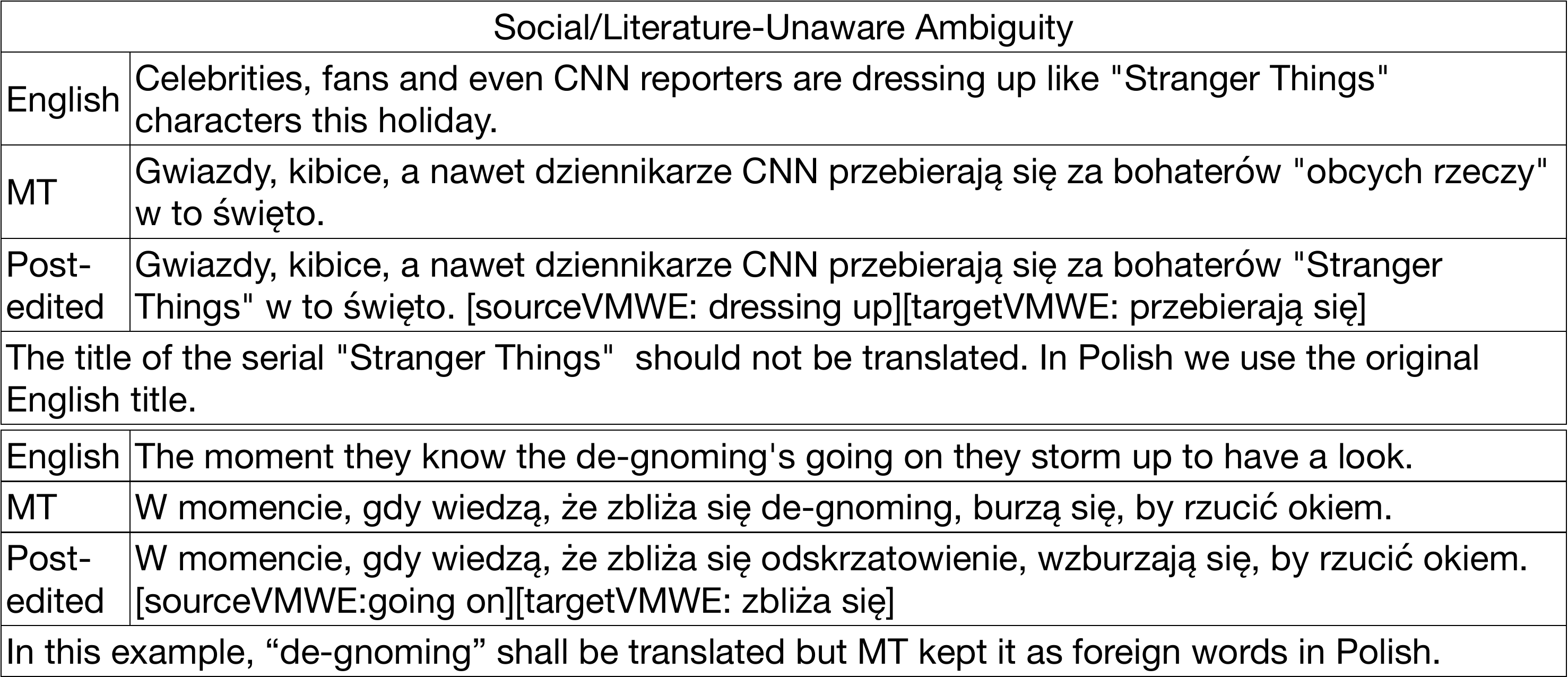}
\caption{Social/Literature-Unaware Ambiguity Example in Polish}
\label{fig:social_liter_un_pl}
\end{center}
\end{figure*}

\begin{figure*}[!h]
\begin{center}
\centering
\includegraphics*[width=\textwidth]{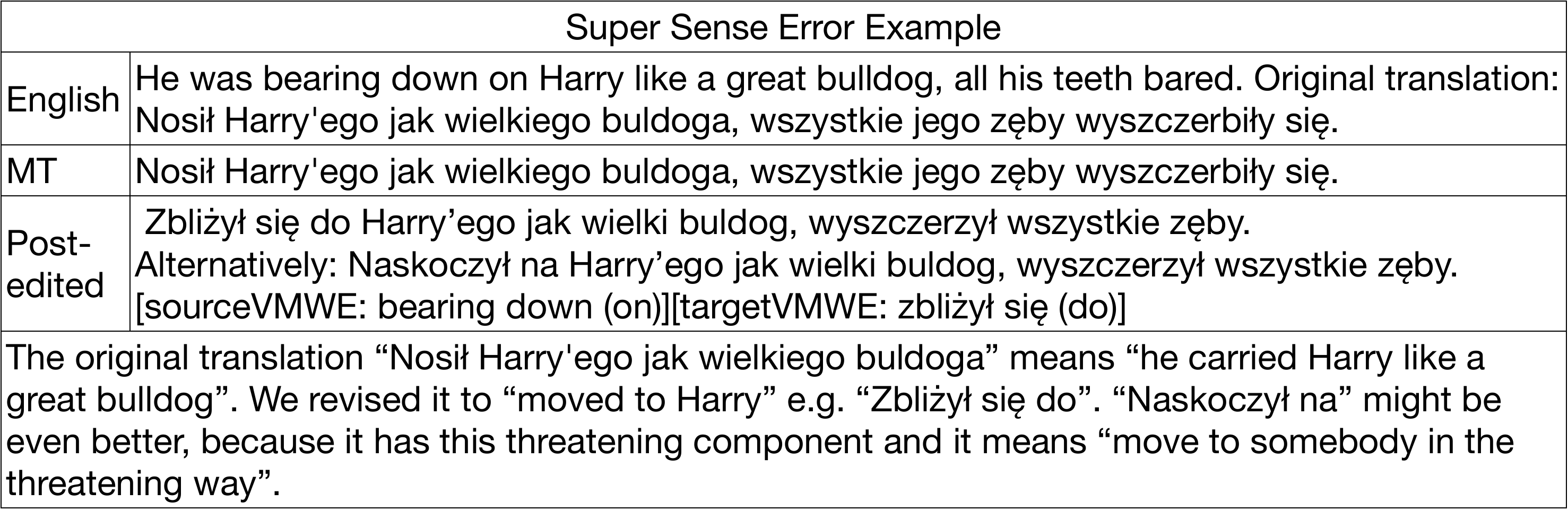}
\caption{Super Sense Error Example in Polish}
\label{fig:super_sense_pl}
\end{center}
\end{figure*}

\begin{figure*}[!h]
\begin{center}
\centering
\includegraphics*[width=\textwidth]{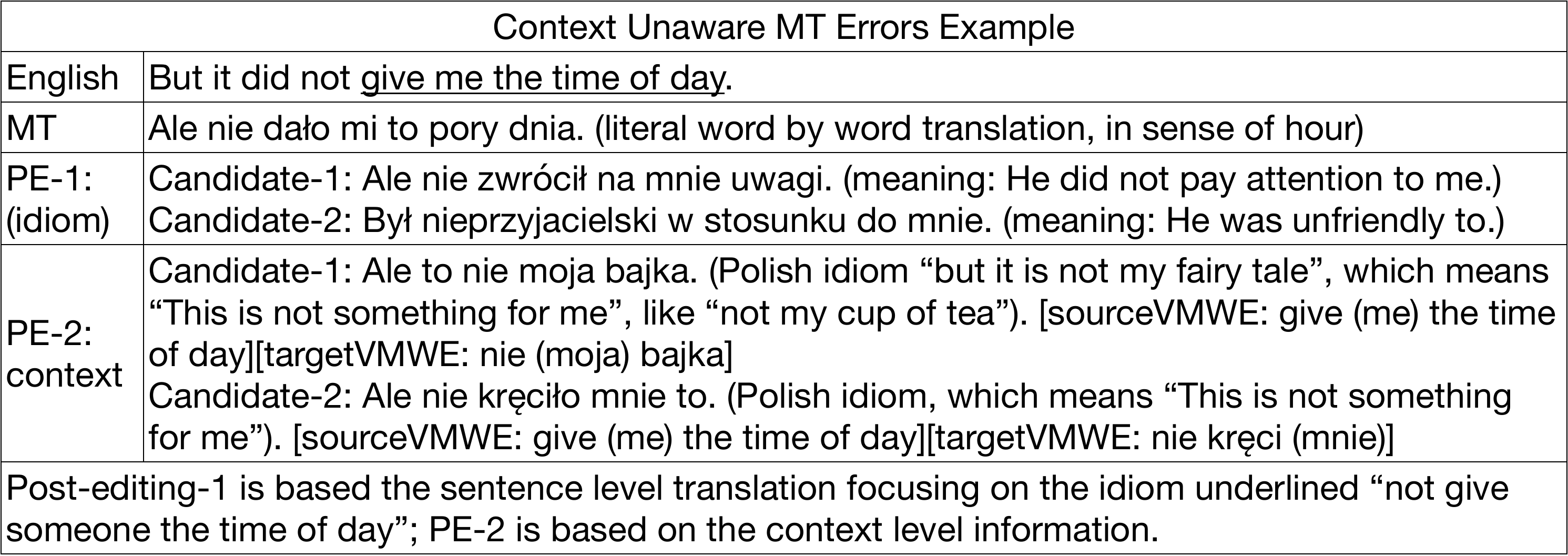}
\caption{Context-Unaware MT Error Example in Polish}
\label{fig:context_unaware_pl}
\end{center}
\end{figure*}

\begin{figure*}[!h]
\begin{center}
\centering
\includegraphics*[width=\textwidth]{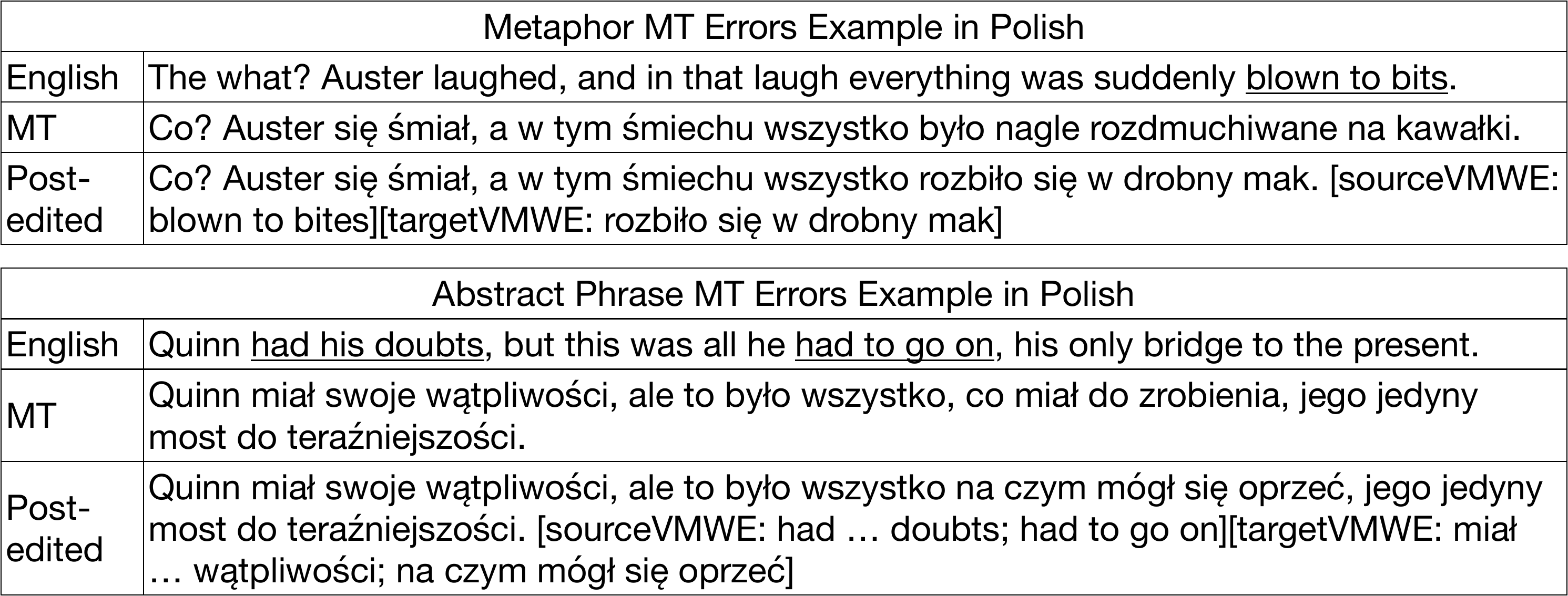}
\caption{Metaphor and Abstract Phrase MT Error Examples in Polish}
\label{fig:metaphor_abstract_phrase_pl}
\end{center}
\end{figure*}

\begin{figure*}[!h]
\begin{center}
\centering
\includegraphics*[width=\textwidth]{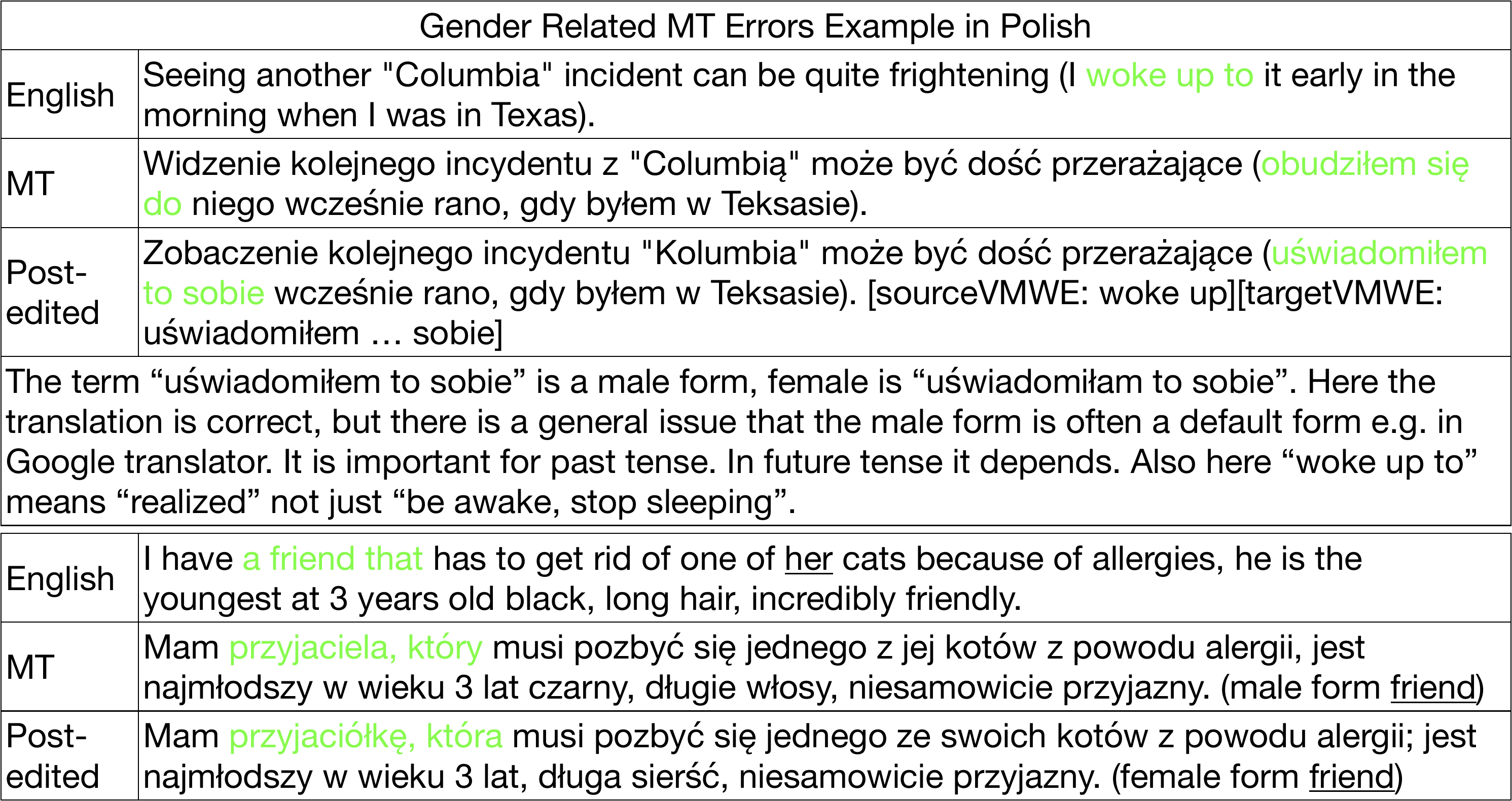}
\caption{Gender Related MT Error Example in Polish}
\label{fig:gender_error_pl}
\end{center}
\end{figure*}
%



\end{document}